\pdfoutput=1

\documentclass{new_tlp}
\usepackage{mathptmx}

\makeatletter
\let\O@argtabularcr\@argtabularcr
\def\O@xtabularcr{\@ifnextchar[\O@argtabularcr{\ifnum 0=`{\fi}\cr}}
\let\O@tabacol\@tabacol
\let\O@tabclassiv\@tabclassiv
\let\O@tabclassz\@tabclassz
\let\O@tabarray\@tabarray
\def\author@tabular{\authorsize\def\@halignto{}\@authortable}
\let\endauthor@tabular=\endtabular
\def\author@tabcrone{{\ifnum0=`}\fi\O@xtabularcr\affilsize\itshape
 \let\\=\author@tabcrtwo\ignorespaces}
\def\author@tabcrtwo{{\ifnum0=`}\fi\O@xtabularcr[-3\p@]\affilsize\itshape
 \let\\=\author@tabcrtwo\ignorespaces}
\def\@authortable{\leavevmode \hbox \bgroup $\let\@acol\O@tabacol
 \let\@classz\O@tabclassz \let\@classiv\O@tabclassiv
 \let\\=\author@tabcrone \ignorespaces \O@tabarray}
\makeatother

\usepackage{epsfig}
\usepackage{amssymb}
\usepackage{theorem}
\usepackage{url}
\usepackage{verbatim}
\usepackage{array}
\usepackage{subfigure}
\usepackage{color,soul}
\usepackage{courier}
\usepackage{afterpage}
\usepackage{adjustbox}
\usepackage{moreverb}
\usepackage{graphicx,verbatimbox}
\usepackage{booktabs}
\usepackage{listings}

\usepackage[colorlinks=true,%
            linkcolor=blue,%
            citecolor=blue,%
           urlcolor=blue,bookmarksopen,bookmarksnumbered,]{hyperref}
\usepackage{url}

\definecolor{Gray}{gray}{0.85}

\newenvironment{TBF}{\color{black}}{\color{black}}

\title{\bf OntoScene, A Logic-based Scene Interpreter: \\Implementation and Application\\ in the Rock Art Domain}
\shorttitle{OntoScene: a Logic-based Scene Interpreter}

\author[Daniela Briola, Viviana Mascardi, and Massimiliano Gioseffi]
         {\centerline{\mbox{DANIELA BRIOLA}}\\
           \centerline{\mbox{Department of Computer Sciences, Systems and Communications}}\\ 
           \centerline{\mbox{University of Milano Bicocca, Italy}}\\
           \centerline{\mbox{\email{daniela.briola@unimib.it}}}\\
           \and  
           \centerline{\mbox{VIVIANA MASCARDI and MASSIMILIANO GIOSEFFI}}\\
           \centerline{\mbox{Department of Informatics, Bioengineering, Robotics, and Systems Engineering}}\\ 
           \centerline{\mbox{University of Genova, Italy}}\\
           \centerline{\mbox{\email{viviana.mascardi@unige.it, gmaxsun89@gmail.com}}}}

\begin{document}

\maketitle

\begin{abstract}
We present OntoScene, a framework aimed at understanding the semantics of visual scenes starting from the semantics of their elements and the spatial relations holding between them. OntoScene exploits ontologies for representing knowledge and Prolog for specifying the interpretation rules that domain experts may adopt, and for implementing the SceneInterpreter engine. Ontologies allow the designer to formalize the domain in a reusable way, and make the system modular and interoperable with existing multiagent systems, while Prolog provides a solid basis to define complex rules of interpretation in a way that can be affordable even
for people with no background in Computational Logics.
The domain selected for experimenting OntoScene is that of prehistoric rock art, which provides us with a fascinating and challenging testbed.

{\it Under consideration in Theory and Practice of Logic Programming (TPLP)}
\end{abstract}


\section{Introduction}\label{sec:intro}

Human perception of complex visual scenes has been studied for a long time in psychology and neuroscience \cite{KondoHirohitoM2017Aavs}: according to the seminal work on ``high-level scene perception'' \cite{henderson1999high}, besides  low-level or early vision, concerned with extraction of
physical properties such as depth, color, and texture from an image \cite{marr1982}, and intermediate-level vision, concerned with extraction of shape and spatial relations
that can be determined without regard to meaning \cite{ullman1996}, a further level of vision is required to perceive and understand a scene:
\begin{quote} high-level vision concerns
the mapping from visual representations to meaning and includes [...] the identification of objects and scenes.
\end{quote}
\begin{TBF}In their recent studies, Kveraga, Bar, and Baldassano \cite{10.2307/j.ctt9qf9vg,baldassano2015}\end{TBF} demonstrate that the brain has regions related to higher-order properties like overall geometry, interactions between objects, aesthetic beauty or memorability of a scene. These regions show a larger response to full scenes than to isolated objects.

Artificial Intelligence can play a major role in modelling and understanding, on the one hand, and reproducing, on the other, the way visual scenes are interpreted by humans. 

While deep learning has shown impressive potential in recognizing images \cite{he2016deep,DBLP:journals/corr/SimonyanZ14a,wan2014deep,donahue2014decaf}, hence providing an ideal tool for low-level and intermediate-level vision,  tackling the high-level vision and associating a meaning with complex scenes may require an explicit and symbolic representation of the domain knowledge, and the ability to reason over it.
%, and the ability to coordinate many different and complex tasks needed for achieving the final goal. 

To understand the semantics of a scene starting from the semantics of its elements and the relations holding among them %, be them binary spatial ones, or n-ary ones involving more elements, in this paper 
we developed OntoScene, which exploits a powerful combination of ontologies and Prolog: ontologies are used for representing knowledge, and Prolog for specifying the rules that domain experts actually use to interpret visual scenes and for implementing the scene interpreter engine.
OntoScene  also relies upon technologies developed in the multiagent systems (MASs) area: it is in fact part of a holonic MAS \cite{holonicMas} named IndianaMAS \cite{ICIC14,euromed,BriolaIDC16,briola2017agent} where agents and MASs devoted to multilingual text understanding, hand-drawn sketch recognition, human interaction, and integration of digital libraries, cooperate and coordinate with the OntoScene framework to classify heterogeneous digital objects.

Following a widely accepted approach for the interpretation of a scene, we consider a scene as an instance or phrase of a visual language where, by analogy with textual languages, relevant graphical symbols can be understood as lexical components or tokens that can be aggregated through the syntactic rules defined according to relations holding among them. Tokens are the sub-images that make up the scene, the grammar is represented by rules defined by the domain expert, and  geometric relationships are ``vertical'', ``overlapping'', ``close'', and the like, and represent aggregation operators.
To allow domain experts to describe the rules for interpreting scenes using a language close to the one in which these rules would be expressed in natural language, we use Prolog. We have designed a user-friendly language that domain experts may use. This rule-based, domain specific language is very similar to Prolog but it hides most Prolog technicalities and can be compiled into standard Prolog clauses. 

OntoScene consists of:
\begin{itemize}
\item Detector and Classifier, two external modules (whose functioning is outside the scope of this paper, and which could be based on our own previous proposals \cite{briola2017agent} or on more recent deep learning techniques) that partition the input image into tokens and associate a list of classifications with them, respectively;
\item SceneInterpreter, the Prolog core of OntoScene; it reasons on a symbolic representation of images that make up a scene and returns their interpretations;
\item OntoScene Agent, an agent providing the interface between OntoScene and the other agents in IndianaMAS;
\item the OntoScene Ontology, which models general concepts needed by OntoScene to work, as well as domain-dependent concepts.
\end{itemize}

To show the potentiality of the OntoScene framework, and to verify the concrete applicability of the proposed solution, we exploit it for the interpretation of complex scenes from the rock art domain, in particular the one of Mount Bego, in France: Mount Bego archaeological site is well-known for its petroglyphs (carvings on rocks), ancient testimonies of human first activities \cite{Bianchi,Bicknell13,Lumley,Bego}. These carvings represent animals, geometric shapes, rural elements and anthropomorphic figures, often represented together to form complex scenes: if identifying and interpreting single elements could be quite simple,
%once the historical period has been studied, 
interpreting complex scenes requires a very detailed knowledge of the domain and offers a challenging testbed to OntoScene.

The core functionality of SceneInterpreter, namely the generation of all the possible scene interpretations according to the interpretation rules, is implemented by Donald Knuth's {\em Algorithm X} for the exact cover problem \cite{DancingLinks00}. 
Algorithm X is a state space searching algorithm that natively exploits depth-first search and backtracking: Prolog turns out to be the perfect language for its implementation. Also, Prolog is very effective as a scene interpretation rule modelling language. Such rules are either sketched by the domain experts using the user-friendly syntax that we devised to mask Prolog details, or written by ourselves in close cooperation with the experts: in both cases, the domain expert that we involved in the experiments, the archaeologist Dr. Nicoletta Bianchi, easily grasped the concepts of unification and backtracking, that allowed her to specify the rules she had in mind, often based on a generate and test technique, in a natural and intuitive way.

The paper is organized as follows: Section \ref{sec:backRel} offers the background knowledge needed for reading the paper and overviews works related to ours; Section \ref{sec:OntoScene_Framework} provides a gentle introduction to OntoScene; Section \ref{sec:ontoandgeom} describes how we modelled domain and spatial knowledge; Section \ref{sec:SceneInterpreter} presents the SceneInterpreter module and exemplifies its functioning on a synthetic domain; Section \ref{sec:CaseStudy} describes the  experiments carried out in the rock art domain; Section \ref{sec:concl-future} concludes and outlines the future directions of our research.

% !TEX root =  ms.tex

\section{Background and Related Work}\label{sec:backRel}

OntoScene is used inside the IndianaMAS holonic multiagent system, which has been designed and developed as a JADE (Java Agent DEvelopment Framework \cite{bellifemine2007developing}) MAS. Although OntoScene's main components are not agents, its interface towards the IndianaMAS components is the JADE \texttt{OntoScene Agent}, which heavily exploits the tools that JADE offers to integrate ontologies in the MAS. Assuming the reader is familiar with knowledge representation in general and with ontologies in particular\footnote{The reader may find an introduction to computational ontologies in \cite{guarino2009ontology}, the specification of the OWL Web Ontology Language in \cite{mcguinness2004owl}, and information on Prot\'eg\'e on the official website, \url{https://protege.stanford.edu/}, accessed on July 2019.}, in Section \ref{sec:background} we provide a brief introduction to IndianaMAS, to JADE, and to the way ontologies are supported therein. We also provide references to the JPL Library\footnote{\url{http://www.swi-prolog.org/packages/jpl/},  accessed on July 2019.} for interfacing SWI-Prolog and Java, and to the JTS Topology Suite\footnote{\url{https://locationtech.github.io/jts/},  accessed on July 2019.} we used to compute relationships among elements in a scene.  
Section \ref{sec:related}  compares our work with related proposals in the logic-based visual languages field, and with spatial ontologies and ontology-driven scene interpretation.

\subsection{Background}\label{sec:background}

\paragraph{IndianaMAS.}

According to the seminal paper by Michael Wooldridge and Nicholas R. Jennings \cite{DBLP:journals/ker/WooldridgeJ95}, an agent is a hardware or, more usually, a software-based computer system that is autonomous (agents operate without the direct intervention of humans or others, and have some kind of control over their actions and internal state); social (agents interact with other agents and possibly humans via some kind of
agent communication language); situated and reactive (agents perceive their environment and respond in a timely fashion to changes that occur in it); and pro-active (agents do not simply act in response to their environment, but they are able to exhibit goal-directed behaviour by taking the initiative). 

Agents are the right tool for coordinating the functioning of software artifacts that show different capabilities and are possibly distributed across a network, with the purpose of making the software architecture as modular, flexible, and reliable as possible. 

The ``Indiana MAS and the Digital Preservation of Rock Carvings: A Multi-Agent System for Drawing and Natural Language Understanding Aimed at Preserving Rock Carvings'' project (``IndianaMAS'' for short\footnote{We use ``IndianaMAS'' to denote both the funded project and the MAS that resulted from it. The project web site, \url{http://indianamas.disi.unige.it/}, accessed on July 2019, gives access to all the project's deliverables and papers.}), funded by the Italian Ministry for Education, University and Research, MIUR, and spanning from March 2012 to February 2015, developed a technology platform based on intelligent software agents for the digital preservation of rock carvings, which both integrates and complements the techniques usually adopted to preserve heritage sites. 
IndianaMAS enables the preservation of all kinds of available data about rock carvings, such as images, geographical objects, textual
descriptions of the represented subjects, allowing the domain experts to organize and structure such digital objects in a standard way and to supply domain experts with facilities for issuing complex queries on the data repositories.

The choice of agent technology for addressing the IndianaMAS goals was a very natural one, given the need that each component of the system, while operating in a highly autonomous way, could interact and coordinate with the other components to share information and to reason about it in the most effective way. 
As discussed \begin{TBF}by Mascardi at al.\end{TBF} in \cite{ICIC14}, the three key services offered by IndianaMAS (sketch recognition, image recognition, and multilingual access to digital libraries) are provided by systems that may be MASs themselves, and that are seen as black boxes by the IndianaMAS agents.

\begin{figure}[h]
\begin{center}
\includegraphics[width=11cm]{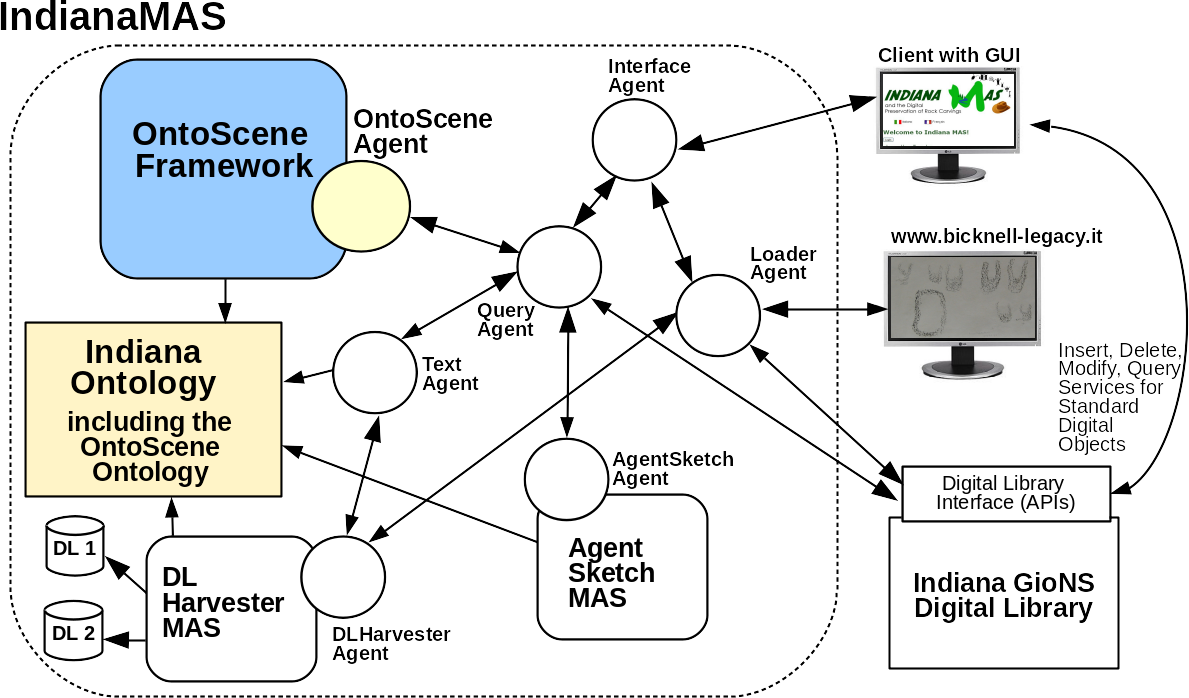} 
\caption{IndianaMAS architecture and data flow.}
\label{fig:indianaMAS}
\end{center}
\end{figure}

Besides OntoScene, the main components of IndianaMAS, sketched in Figure \ref{fig:indianaMAS}, are:
 
% \begin{itemize}
% \item The Indiana Ontology, which structures the domain of interest; it consists of sub-ontologies, among which the OntoScene Ontology\footnote{The OntoScene Ontology shares the ``Thing'' class with the Indiana Ontology: we will not address the question of whether it should be named an ``ontology'' or a ``sub-ontology'', as this distinction is not relevant for the paper, and we will always use ``ontology'' to describe it.}, and is accessed by all the agents and components in the system.
% %
% \item AgentSketch, which interprets manual drawings based on the Indiana Ontology.
% %
% \item The Indiana GioNS Digital Library, which contains all the digital objects inserted into the system by registered users, together with their metadata, needed for their later retrieval.
% %
% \item MANENT, whose main purpose is accessing digital libraries spread around the world, besides the Indiana GioNS one, and to classify multilingual documents available there by applying natural language processing techniques both to the document textual metadata and to the text body. 
% %
% \item Client with a graphical user interface, for interacting with IndianaMAS.
% %
% \item Presentation Agent, for suitably presenting the content to the users depending on their profiles, permissions and device.
% %
% \item User Manager Agent, for personalization and profiling purposes.
% %
% \item Insert-Modify-Delete Manager and Query Manager Agents, for preparing the right queries and requests to the Indiana GioNS digital library. 
% \end{itemize}

\begin{itemize}
\item The Indiana Ontology, which structures the domain of interest; it consists of sub-ontologies, among which the OntoScene Ontology\footnote{The OntoScene Ontology shares the ``Thing'' class with the Indiana Ontology: we will not address the question of whether it should be named an ``ontology'' or a ``sub-ontology'', as this distinction is not relevant for the paper, and we will always use ``ontology'' to describe it.}, and is accessed by all the agents and components in the system.

\item Client with a graphical user interface, for interacting with IndianaMAS.

\item The Indiana GioNS Digital Library, which contains all the digital objects inserted into the system by registered users, together with their metadata, needed for their later retrieval.

\item Text Agent, able to interpret multilingual documents according to the Indiana Ontology.

\item Query Agents, each managing one query coming from the client.

\item Loader Agent, collecting new data from external resources like the Bicknell Legacy web site\footnote{\url{http://www.bicknell-legacy.it}, accessed on July 2019.} and managing the creation and insertion of new digital objects into the Indiana GioNS Digital Libray.

\item Interface Agent managing the creation of new Query Agents.

\item The Digital Library (DL) Harvester MAS, which independently and proactively searches digital libraries on the web to retrieve new images and texts related to the domain modelled by the Indiana Ontology.

\item AgentSketch MAS, which interprets manual drawings based on the Indiana Ontology.
\end{itemize}

\paragraph{JADE.}
JADE is a Java-based software platform that supports the development of agents and MASs thanks to a Graphical User Interface and tools supporting the MAS debugging and deployment phases. JADE MASs can be distributed across machines in a way that is fully transparent to the developer. The minimal system requirement is the Java runtime environment or JDK, version 5.

\paragraph{Ontologies in JADE.}
JADE helps developers in achieving semantic interoperability between agents thanks to a simple and fast way to exploit ontologies directly inside the platform and the agents: agents can exchange messages referring to a shared ontology, and then rely on the JADE Ontology management offered by the ContentManager class.
The developer may use an ontology to formalize what the agents knows (Concepts and Predicates) and can do (Actions), and share this ontology among the agents: in this way, knowledge is modelled outside the agents, boosting modularity and reuse, and the content of messages is based on a shared ontology, facilitating interactions and simplifying the serialization phase that is then demanded to the JADE platform.

The three types of objects considered when creating an ontology for JADE are:
\begin{itemize}
\item Predicates: boolean expressions describing something about the agent environment or its beliefs.
\item Concepts: structured objects describing the elements of the world, and their relationships.
\item Agent actions: special Concepts modeling what an agent can do and can be requested to do with a message.
\end{itemize}

If two agents share the same ontology, one agent can request the other to perform an Action, and can receive an answer containing the Action results, which will be a Concept, a Predicate, or a list of them.

%Manually Creating this types of Java representation of an ontology (which is usually already written in an ontology language) can be done, but is complex and long above all if the ontology is complex: furthermore, every time the ontology changes, its Java representation must be updated, and above all, often developers are not the same people who design the ontology, and both the involved figures are usually not abel to write both in OWL (or similar language) and in Java. Consequently, designers and expert of the domains should focus on modeling the ontology in a language, and developers should focus on managing this ontology in Java for Jade. 

To allow developers to automatically generate a Java  representation  of  an OWL  ontology  coherent  with the JADE requirements, the tool OntologyBeanGenerator\footnote{\url{https://protegewiki.stanford.edu/wiki/OntologyBeanGenerator}, accessed on July 2019.} can be adopted. The latest version of OntologyBeanGenerator available on the official website is 4.1, including a basic ontology modeling the over mentioned concepts: domain specific concepts must be added as subclasses of Concept, Predicate and Agent, and then the tool will provide a Java representation of the ontology, directly usable by JADE.

Given some limitations of that version, we developed OntologyBeanGenerator 5.0 \cite{WOA18} as a new Prot\'eg\'e plugin\footnote{We asked and obtained a written consent from the author of OntologyBeanGenerator, Chris van Aart, to extend the original source code.}.
%
%\paragraph{OntologyBeanGenerator 5.0.}
%\label{sec:BeanGenerator2}
%
%To support the designer in the automatic exportation of an OWL ontology into a Java format that JADE is directly able to exploit, we developed the OntologyBeanGenerator 5.0 . 
OntologyBeanGenerator 5.0 (OBG5.0 in the sequel, available from \url{www.disi.unige.it/person/MascardiV/Download/OBG5.0.zip}) has been developed with three goals in mind: correcting some bugs of OntologyBeanGenerator 4.1; adding the methods and exceptions management directly inside the ontology; and producing an additional output to support the OntoScene framework.

The main improvements of OBG5.0 w.r.t. OntologyBeanGenerator 4.1 are:
\begin{itemize}
\item Addition of a new tab called Java Method Mapper to manage the methods creation and exportation: the purpose of the new tab is to offer the designer of the ontology a way to directly add methods to the Java version of the ontology: in the previous version, the only option was to add a property and consequently to get the setter and getter automatically. 
%This is mainly foreseen if we want to add methods that cannot be mapped in a setter/getter of a specific field. %These methods are saved in an external XML file which is loaded by the tab and overwritten when exporting the ontology;
\item Exception management: methods are allowed to raise Exceptions. To do this, a specific ontology to be imported has been created. Thanks to this addition, Exceptions can be exchanged between agents too, since they are a subclass of Concept.
\item Correction of some bugs that were present in the ontology generation stage. %: now the ontology correctly extends BeanOntology;
%\item Automatic import of the Default Implementation of the classes in the Java Ontology;
\item Possibility to export the class hierarchy in a Prolog format: in order to implement Prolog rules that reason about the ontology, we need a Prolog representation of it. 
%Currently, we only need the classes hierarchy. 
To achieve this goal we added an automatic ontology export functionality to OBG5.0. The obtained Prolog representation only formalizes the classes hierarchy, as this is the only knowledge we currently need in OntoScene.
%, to get a Prolog version of what we needed in the framework.
%: this addition does not change in any way the other functionalities of OBJ5.0.

% In the export tab, where the user specifies the name and path for the Java export of the ontology, (s)he can also list the ontology classes to be exported in Prolog, separated by a ``;": these classes, and all their subclasses, will be automatically exported in Prolog in a file named {\it subclasses.pl}. The obtained Prolog representation only formalizes the classes hierarchy, as this is the only knowledge we currently need in OntoScene.

\begin{figure}[!h]
\begin{center}
\includegraphics[width=4cm]{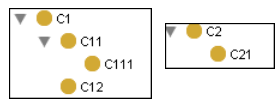}
\caption{A simple ontology to be exported in Prolog.}
\label{fig:prolog}
\end{center}
\end{figure}

As an example, the Prolog version of the class hierarchy shown in Figure \ref{fig:prolog} is:
\begin{verbatim}
subclass_of('C1', 'C11').
subclass_of('C1', 'C12').
subclass_of('C11', 'C111').
subclass_of('C2', 'C21').
\end{verbatim}
\end{itemize}

\paragraph{The JPL Library.} JPL can be used to embed SWI-Prolog in Java as well as for embedding Java in SWI-Prolog. In both setups it provides a  bidirectional interface. The two predicates that we used for accessing Java from inside SceneInterpreter are:

\verb|jpl_new(+X, +Params, -V)|
    where \texttt{X} is an object (non-array) type or descriptor and \texttt{Params} is a list of values or references, unifies \texttt{V} with the result of an invocation of that type's most specifically-typed constructor to whose respective formal parameters the actual \texttt{Params} are assignable (and assigned). 
    
\verb|jpl_call( +V, +Method, +Params, -Result)|
    unifies \texttt{Result} with a JPL reference to (or value of) the result of calling the named \texttt{Method} of \texttt{V} with \texttt{Params}.

\paragraph{The JTS Suite.} 
The JTS Topology Suite (JTS) is an open source Java software library that provides an object model for planar geometry together with a set of fundamental geometric functions. In OntoScene, it was used to implement the basic relations between regions that characterize the Region Connection Calculus (RCC, \cite{DBLP:journals/ai/LiY03,DBLP:conf/kr/RandellC89}) such as \texttt{disjoint}, named ``Disconnetted'' in RCC, \texttt{overlap} (``Partially Overlapping'' in RCC), and \texttt{contains} (``Non-Tangential Proper Part Inverse'' in RCC), plus further derived relations.

\paragraph{Using JPL and JTS together.} If \texttt{GR} is a reference to the implementation of the interface for geometric relations, 
\verb|jpl_call(GR, contains, [BB1, BB2], @(true))| succeeds if the method \texttt{contains}, implemented in Java by exploiting the API offered by JTS, and called on \texttt{BB1} (where BB stands for the ``Bounding Box'' of one image) and \texttt{BB2} (the bounding box of another image), returns \texttt{true}. 
In a similar way, we may have the following predicate calls in a Prolog piece of code
\begin{itemize}
\item \verb|jpl_call(GR, overlap, [BB1, BB2], @(true)))|, 
\item \verb|jpl_call(GR, vertical, [BB2, BB1, 'up'], @(true))|, 
\item \verb|jpl_call(GR, near, [BB1, BB2, 0.5], @(true))| (\texttt{0.5} is the threshold for considering two bounding boxes close to each other, expressed in pixels),
\item ...
\end{itemize}
with their intuitive meaning, better explained in Section \ref{subsec:geomrel}.
As a more complex example, the call to \verb|jpl_call(GR, group, [JavaBBs, 0.5], @(true))| works if \texttt{JavaBBs} is the Java representation of a Prolog list, and the \texttt{group} method called on that list with \texttt{0.5} as proximity threshold returns true.

\subsection{Related work}\label{sec:related}

\paragraph{Logic-based visual languages.}

Many approaches for dealing with visual languages have been proposed in the literature: this research area has a long tradition, with both an ad-hoc conference established in 1984, VL/HCC\footnote{The most recent edition of VL/HCC dates back to 2018, \url{https://vlhcc18.github.io/index.html}, accessed on July 2019.}, and a high quality journal, the Journal of Visual Languages and Computing.\footnote{\url{https://www.journals.elsevier.com/journal-of-visual-languages-and-computing}, accessed on July 2019.}
In this section we review some approaches that use logical or relational formalisms for recognizing and understanding visual languages, starting from the older and more established ones, and moving towards more recent proposals. A complementary approach, which is out of the scope of this paper, is to use visual programming approaches to specify logic-based languages\begin{TBF}, as done by Ladret and Rueher \cite{ladret1991vlp} and Agust\'i et al. \cite{agusti1998visual}\end{TBF} %\cite{ladret1991vlp,agusti1998visual}. 

Defining visual languages using a logic-based language in general, and Prolog in particular, ensures that declarative and operational semantics can be shared among humans and between humans and machines. The declarative semantics allows both humans and machines to reason about the specification independently of the implementation, while the operational semantics allows the generation and recognition of images defined by the specification. After a very active period in the early nineties of the last century, the ``logic-based visual languages'' research field has produced less results, probably due to the raise of statistical approaches in the meanwhile. 

\begin{TBF}Crimi et al. introduced the concept of relational grammars \cite{crimi1991relation}: while\end{TBF} textual languages use an implicit sequential concatenation relationship, the proposed extension relaxes this constraint by providing an arbitrary number of geometric relationships.
\begin{TBF}Helm and Marriott defined the relationships between images and their meaning via a class of declarative and constraint-based specification languages, written in Prolog \cite{helm1991declarative}, and Wittenburg et al. presented a formalism called {\em unification-based grammar} and a parsing algorithm for visual languages in \cite{wittenburg1991unification} \end{TBF}. The formalism extends D-PATR \begin{TBF} \cite{karttunen1986d}\end{TBF} with logical constraints and a new  bottom-up parsing method.
\begin{TBF}Meyer introduced a new technique to extend Logic Programming with terms representing partially specified images \cite{meyer1992pictures}\end{TBF}. To this aim, the Picture Clause Grammar (PCGs), a form of specification for visual languages ​​similar to the Definite Clause Grammar (DCGs) of textual languages, is defined. None of these proposals come with an implemented prototype, making their practical applicability limited.

\begin{TBF}Santosh et al.'s proposal \cite{5277729}\end{TBF} is close to ours both in the system architecture and in the methodological approach, but not in the final goal. They aim at expressing graphic
symbols by a number of graphical primitives that may be of any complexity and connecting relationships that can be deduced from state-of-the art image treatment and analysis tools. The existence of suitable tools for image pre-processing is also assumed by us, by including the Detector and Classifier modules presented in the next sections in the OntoScene architecture. 
The symbolic representation obtained by the image analysis tools is then provided to an inductive logic programming solver that outputs a set of logical rules that define the positive example
set. On the contrary, we provide the symbolic representation of elements detected in the scene to a Prolog program that, thanks to rules that model the domain expert knowledge, provide a semantic interpretation of the scene. 

\begin{TBF}Antanas et al. present\end{TBF} a framework combining compositional hierarchies,
qualitative spatial relations, relational instance-based learning and robust feature extraction \cite{antanas2012relational}.
For each layer in the hierarchy, sub-structures in the images are detected, classified and then employed one layer
up the hierarchy to obtain higher-level semantic structures, by making use of qualitative spatial relations implemented in Prolog. 
%The approach is applied to street view images. The implementation combines code written in Prolog,  Matlab and C, where Prolog is used to represent spatial relations. 
Given that we may have scenes that include scenes, we support a hierarchical structure as well. So far, we only employed two levels in the hierarchy (one scene that includes another scene, that only includes ``atomic'' tokens, as in Table \ref{tab:ritSac}, third and fourth images) but there are in principle no reasons for adding more layers. 
W.r.t. that work, we also have a domain ontology and a MAS coordinating the interactions among the framework components.  

\begin{TBF}In their work, Di Martino and Esposito do not consider any low-level image processing stage\end{TBF}, but integrate a domain ontology in the system architecture, like in OntoScene \cite{doi:10.1002/spe.2336}: the authors describe a procedure and a prototype implementation for the automatic recognition of design patterns from documentation of software artifacts design and implementation, provided in XMI\footnote{The XML Metadata Interchange (XMI), \url{https://www.omg.org/spec/XMI}, accessed on July 2019.}. The procedure exploits a semantic representation of the patterns to be recognized, based on an existing ontology. Both the UML set of diagrams related to the analysed software artifacts and the patterns represented in the ontology are translated into a Prolog knowledge base. A Prolog program implements the heuristics and features that trigger the recognition on that knowledge base.

Although not based on logic programming, it is worth mentioning \begin{TBF} the work by Hammond and Davis \cite{Hammond:2007:LSL:1281500.1281546}, which uses the rule-based language Jess \cite{hill2003jess} for specifying how sketched diagrams in a domain are drawn, displayed, and edited,  and the work by Costagliola et al. \cite{1575738}, which uses \end{TBF} rules named ``sketch patterns'' for describing and recognizing diagrammatic sketch languages, and that are very close to Jess rules. 

%In \cite{wittenburg1993adventures} it is noted that many of the approaches used to define grammars have limitations such as: they do not support an arbitrary sort of input, but allow only some special relationships, require graphs connected, do not provide bottom-up parsing algorithms and do not accept grammars ambiguous. It is highlighted that for some domains all these limitations are not acceptable and are proposed methods of predictive parsing for visual languages ​​whose relationships involved are not necessarily ordered (symmetric and non-transitive).
% In \cite{marriott1994constraint} develops a high-level framework for the definition of visual languages ​​through a new formalism that uses constraints multiset grammars (based on the ideas presented in \cite{helm1991declarative}).
% In \cite{marriott1997classification} (expanding \cite{marriott1994constraint}) a complete hierarchy of visual languages ​​is proposed based on their formal properties (similarly to the Chomsky hierarchy of textual languages) and the expressiveness and cost of the parsing of the classes of identified visual languages ​​(often they are context-sensitive).

\paragraph{Spatial ontologies and ontology-driven scene interpretation.} Research on modeling either spatial or domain-dependent concepts (or both) in an ontology, and exploiting such an ontology for interpreting a graphical scene, is closely connected with our work. 
\begin{TBF}Haarslev et al. present one of the first works in this area \cite{haarslev1994combining}, introducing \end{TBF} ``spatio-terminological inferences'' to mean a three-level view of inference processes combining quantitative, qualitative and conceptual representations. They use the TBox and ABox of LOOM \cite{baader1991jurgen}, and apply spatio-terminological reasoning to parsing visual programming languages. 
Other works by the same research group use different ontology languages and address different application domains, but remain consistent with the seminal proposal. As an example, \begin{TBF}Haarslev et al. exploit description logic and apply ontological reasoning to sketch-based queries for Geographical Information Systems \cite{haarslev1999logic,haarslev2002visual}\end{TBF}. 

\begin{TBF}In his recent book ``Description Logics in Multimedia Reasoning'', Sikos presents\end{TBF} an integrated and comprehensive analysis of issues relevant to our work, with chapters on spatial Description Logics, spatial annotations, and reasoning tools \cite{DBLP:books/sp/Sikos17}. 

\begin{TBF}Forestier et al. and  Bannour and Hudelot present other ontologies for modeling spatial concepts and reasoning on scenes and images \cite{forestier2008evolutionary,bannour2011towards}\end{TBF}. To make a recent example, \begin{TBF}Gu\'erin et al. \cite{Guerin:2017:OFA:3050464.3050589} exploit\end{TBF} one ontology that formalizes the basic concepts of the image processing domain and provides a way
to organize and use input and output data in a formal structure, and provide a formal ontological implementation of the comic books domain. This ontology is meant to handle the content of a
comic book, to support the automatic extraction of its visual components, and to formalize the semantics of the domain's codes.

Whilst taking inspiration from works on spatial ontologies, OntoScene needs to model notions like ``Classification'' and ``Interpretation'' that allow us to distinguish between the ``syntax'' of the image, dealt with by the Detector and Classifier modules, and its semantics, devised thanks to ontological reasoning on the domain, along with logical reasoning. Being a JADE MAS, our framework requires the OntoScene ontology to be compliant with the JADE requirements for ontology management. For these reasons, we could not reuse existing ontologies as they are; moreover, some ontologies were not available to the research community and others were not modeled in OWL,  as needed in our work. Nevertheless, we took them into account when modeling the ``GeometricRelations'' concept.

\section{The OntoScene Framework: a Gentle Introduction}
\label{sec:OntoScene_Framework}

\subsection{The Initial Scenario}
\label{initscen}

\begin{figure}[h]
\begin{center}
\includegraphics[width=4cm]{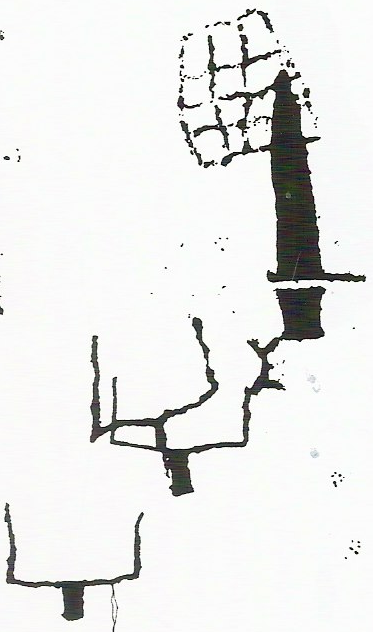} 
\caption{A Complex Scene from the book by de Lumley and Echassoux ({\color{blue}{de Lumley and Echassoux 2009}}), page 176, figure 142. This and all the other images taken from that book are reproduced by kind permission of Professors de Lumley and Echassoux.}
\label{fig:complexscene-no-BB}
\end{center}
\end{figure}

Viviana is very curious about the prehistoric rock art of Mount Bego and she would like to know how a domain expert would interpret the image shown in Figure \ref{fig:complexscene-no-BB} according to the most recent archaeological findings.

In that image, Viviana can only see a ``matrix'' in the top right corner, with a kind of filled trapeze overlapping it, and three symbols very similar to each other, made of lines with filled rectangles in the middle, in the center and the bottom left corner of the image.  

Massimiliano, who is good at detecting and classifying symbols from a purely syntactic point of view, explains her that the ``matrix'' can be classified as a ``Reticulum\_Class'' with 100\% confidence, the trapeze along with the rectangle just below it can be classified as a ``Dagger\_Class'', and the three symbols made of lines with small filled rectangles in the middle, can be classified as ``Up\_Corn\_Class''. These classes are drawn from an ontology modelling information about Mount Bego's petroglyphs. 

Viviana is far from being satisfied, since this syntactic classification says nothing about the meaning of symbols and of the scene as a whole. She sends the information provided by Massimiliano to Daniela, who knows many archaeologists, and asks her if she can provide a semantic interpretation of the scene.  

Daniela contacts Annie and Henry: Annie is very good in associating domain-dependent meaning to symbol classifications. By exploiting the same ontology used by Massimiliano, she can confirm that a symbol classified as a ``Reticulum\_Class'', when interpreted inside a rock art artifact from Mount Bego, actually represents a ``Reticulum'';  in another domain, the ``Reticulum\_Class'' might have been interpreted as ``Prison Bars'' or ``Chess Board'': decoupling the classification from the interpretation fosters reuse and modularity, and the domain ontology is a good means for achieving this aim.  
A ``Dagger\_Class'' represents a ``Dagger'' in the Mount Bego rock art domain, and the ``Up\_Corn\_Class'' represents a ``Corniform''.

The semantics associated by Annie with the classifications devised by Massimiliano is still not enough to interpret  the scene: more knowledge and more reasoning are needed. Taking Annie's interpretation of symbols belonging to the scene into account, Henry reasons about them and their spatial relationships and finally informs Daniela that the dagger and the reticulate at the top of the image identify the ``Storm God'' inside a pastoral scene, characterized by a group of corniforms \cite{Lumley}. Another possible interpretation could be that the two corniforms in the center of the image, one inside the other, identify the ``Bull God'', and the bottom left corniform is a stand-alone symbol, unrelated with the others. However, Henry thinks that the first interpretation is the most likely one. 

Viviana is now happy with this explanation: by moving from symbol classification (symbol syntax) to interpretation (symbol semantics), and then combining interpretations  into coherent sub-scenes via domain-dependent rules, her friends helped her understanding the image. 

\begin{figure}[h]
\begin{center}
\includegraphics[width=9cm]{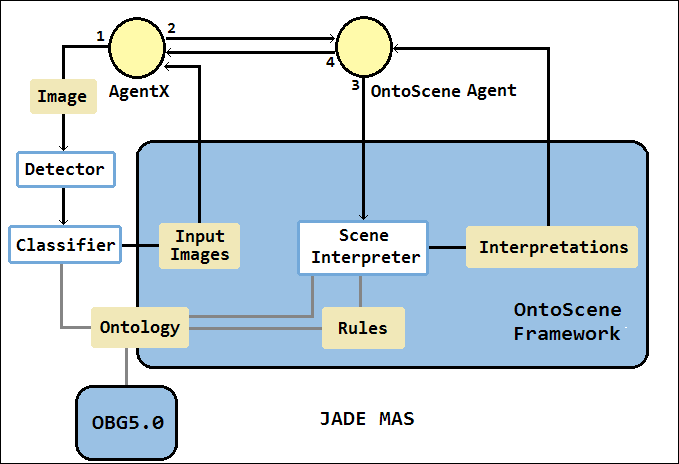}
\caption{OntoScene Framework: architecture and data flow.}
\label{fig:ontoScene1}
\end{center}
\end{figure}

The people involved in this scenario and the way they interact reflect the OntoScene framework that we developed: each person could be suitably associated with an agent or a component in the OntoScene software framework depicted in Figure \ref{fig:ontoScene1}:  
\begin{itemize}
\item Viviana is an unnamed, generic agent \texttt{AgentX} that wants to understand the meaning of a scene depicted inside an input image: she interacts with a software module (Massimiliano) able to detect coherent sub-images, also named ``tokens'', inside an image and to classify them, and with another agent (Daniela) that acts as an interface with the domain experts.
\item Massimiliano plays the role of token \texttt{Detector} and \texttt{Classifier}, and is able to divide an input image into sub-images. The computed set of sub-images, each one associated with a list of possible classifications, is sent back to AgentX, Viviana in this example.
\item Daniela acts as the \texttt{OntoScene Agent}, managing the interactions with Annie and Henry, to provide an interpretation for the image.
\item Annie and Henry implement the intelligent engine able to interpret scenes according to the meaning of classified tokens, and to the rules that aggregate such interpretations (also taking spatial relations into account), to provide a semantics of the complex scenes (\texttt{Scene Interpreter}).
\item All these agents and components share a common ontology.
\end{itemize}

To go deeper inside the high level architecture of OntoScene and the data flow within it, white rectangles in Figure \ref{fig:ontoScene1} represent system modules, while light yellow (light gray in B/W) rectangles represent either data flowing between them, or data that are used by them. Circles represent agents and the blue (dark gray in B/W) rectangles with rounded corners represent the two platforms involved in the process.
%: OBG5.0 has already been described in \cite{WOA18}, while OntoScene is the framework described in this paper.  

An arrow flowing from A to B tagged with data D, represented as a rectangle on the arrow itself, means that D is generated as an output by A and used as an input by B. 
An arrow flowing from A to B with no tag means that A generates some output that becomes an input for B (but we do not need to identify it). 
A gray line between two components means that a ``uses/is used'' relationship holds between them. 

%
%The components needed by OntoScene to interpret an input image are (SECONDO ME SI POTREBBE TOGLIERE):
%\begin{itemize}
%\item Detector and Classifier, two external modules (whose functioning is outside the scope of this paper, already developed and described in \cite{briola2017agent,euromed,BriolaIDC16}) that partition the input image into ``tokens'' and associate a list of classifications with them, respectively (Section \ref{sec:3.2.1});
%\item SceneInterpreter, the core of OntoScene; it captures images (InputImages) that make up a scene and returns their interpretations (Sections \ref{sec:3.2.2} and \ref{sec:SceneInterpreter});
%\item OntoSceneAgent, an agent providing the interface between OntoScene and the other agents in the MAS, identified as AgentX (Section \ref{sec:3.2.4});
%\item OBG5.0, introduced in Section \ref{sec:background}.
%\end{itemize}
%

Data managed by OntoScene are:
\begin{itemize}
\item Image, the raw input image to be interpreted;
\item InputImages, the output of the Detector and Classifier representing the input image and the tokens therein, along with their bounding boxes and their classifications, in a symbolic format;
\item Prolog Rules, which are set by the domain experts and define how to interpret an image; %they must be consistent with the ontology;
\item Interpretations, which represent the final output computed by OntoScene;
\item Ontology, which represents the application domain, namely the classifications, interpretations and geometric relationships that are meaningful for the specific image domain and interpretation task; these concepts are used by the Rules (Section \ref{sec:3.2.3}).
\end{itemize}

\subsection{Syntactic Pre-processing: Detector and Classifier}
\label{sec:3.2.1}

The interpretation of the input scene requires that it has been segmented into atomic sub-images (``tokens'') and that one or more  classifications have been associated with each of them. To this aim, we assume the availability of a \texttt{Detector} and \texttt{Classifier}.

We do not enter into the details of how these modules could be designed and implemented, since many libraries and tools for solving the bounding box detection and the classification problems exist and are available to the community. 
Just to make some examples, the MathWorks Image Processing Toolbox\footnote{\url{https://www.mathworks.com/products/image.html}, accessed on July 2019.}  provides algorithms for image processing, analysis, visualization, segmentation; OpenCV\footnote{\url{http://opencv.org/}, accessed on July 2019.}, cross-platform and free for both academic and commercial use, offers 2D segmentation and recognition functionalities suitable for the implementation of both the Detector and the Classifier, besides many other advanced features; ImageJ\footnote{\url{https://imagej.nih.gov/ij/index.html}, accessed on July 2019.}, written in Java, and Pillow\footnote{\url{https://python-pillow.org/}, accessed on July 2019.}, in Python, are other libraries providing edge detection functionalities useful for implementing the \texttt{Detector} module.

As far as the classification of images in the rock art domain is concerned, we refer to our previous work within the IndianaMAS project, where ad-hoc detection and classification algorithms were developed \cite{briola2017agent,ICIC14}.

\begin{figure}[h]
\begin{center}
\includegraphics[width=8cm]{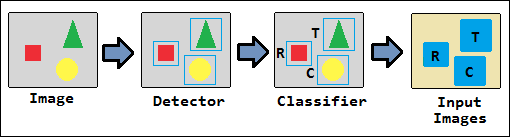}
\caption{How the Detector and Classifier modules interact.}
\label{fig:15}
\end{center}
\end{figure}

To show how the \texttt{Detector} and \texttt{Classifier} modules are expected to work, we consider an example. The input image in Figure \ref{fig:15} contains three figures: a rectangle, a triangle, and a circle. The \texttt{Detector} identifies the three sub-images and associates them with a bounding box rectangle (BB) representing their position and size within the image.
The \texttt{Classifier} analyzes the sub-images identified by the \texttt{Detector} and assigns the R (rectangle), T (triangle) and C (circle) classifications, consistently with the domain ontology.

The \texttt{Classifier} is expected to assign multiple classifications to the detected figures, in case of ambiguity. Its output is hence a list of possible classifications for each BB, with an associated confidence in the interval [0.0, 1.0]. If there are no doubts about the classification, the list will contain one element only.  

% HO TOLTO L'ESEMPIO, mi pare si capisca comunque anche senza

%\begin{figure}[h]
%\begin{center}
%\includegraphics[width=7cm]{images/16.png}
%\caption{Classification output.}
%\label{fig:16}
%\end{center}
%\end{figure}

%Figure \ref{fig:16} shows an (unrealistic) example of ambiguity between a rectangular shape and a circular figure. The shape identified with a question mark might be recognized as R (rectangle) with confidence 0.8 and as C (circle) with confidence 0.2, or -- using a different classification algorithm -- vice versa. The classification of the triangle, not shown in the figure, is instead T with confidence 1.0.
%
%\begin{figure}[h]
%\begin{center}
%\includegraphics[width=5cm]{images/17.png}
%\caption{InputImages: the output of the Classifier, used as input by OntoSceneAgent.}
%\label{fig:17}
%\end{center}
%\end{figure}
%
%
%The output of the Classifier is represented in Figure \ref{fig:17}. If compared with a natural language processing pipeline or with the lexical analysis of a source code in computer science, the classifier acts as a tokenizer: the content of each BB is tagged with its category (or with its possible categories if the classification confidence is not 100\%) that will be afterwards used by the interpreter for understanding the meaning of the scene represented in the input image.  

% 3.2.2.  The interpreter of complex scenes
\subsection{From Syntax to Semantics: SceneInterpreter}
\label{sec:3.2.2}

\begin{figure}[h]
\begin{center}
\includegraphics[width=7cm]{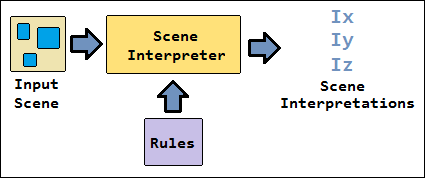}
\caption{The SceneInterpreter module.}
\label{fig:18}
\end{center}
\end{figure}

Figure \ref{fig:18} shows the \texttt{SceneInterpreter}, the core module of OntoScene.
\texttt{Sce\-ne\-In\-ter\-pre\-ter} takes an image consisting of a set of tokens in input (we will call this set a ``scene'')  and returns all its interpretations. It is driven by logical rules that define the possible meanings of each token recognized during the detection and classification stages, and the ``well formed'' scenes that the framework can recognize and interpret along with their meaning. 
%To ensure modularity and reuse, such rules are loaded from a file. In our framework, we adopted Prolog as logic language, so in the remainder of the paper we will show facts and rules directly adopting the Prolog syntax. 

%module can be thought of as a file that SceneInterpreter can read and load. In addition to containing the rules of recognizable scenes, the associations (classification, interpretation) are also known.
%The framework then captures an image (scene) composed of multiple atomic symbols already classified and, using the rules loaded, provides a list of all its possible interpretations. Concepts known in the framework (such as classifications and interpretations, etc.) are part of an ontology (see section 3.2.3).

A figure classified as a \verb|Circle| might be interpreted as a \verb|Planet| in an astronomic domain, as a \verb|Face| in an emoticon recognition domain, as a \verb|Traffic_Light_Element| by a self-driving car: the classification as a circle is not enough to correctly interpret a figure in a context made up of other figures. Making the link between the classification and the interpretation levels explicit allows the designer to reuse the classification output and to change the scene interpretation according to the current domain, by only changing the interpretation rules. 

As an example, in the rock art domain that provides the case study of this work, a figure classified as an \verb|Anthropomorphic_Shape| might be interpreted as a \verb|Human|, a figure classified as a \verb|Line_Shape| might be either a \verb|Sword| or a \verb|Staff|, and a triangle should be interpreted as a mage cap.

The interpretation of an individual token is defined by means of the \texttt{interpretation(Cl, ImgInt)}  fact that associates the interpretation \verb|ImgInt| with the classification \verb|Cl|.
In the rock art example, interpretation facts might look like
{\small{
\begin{verbatim}
interpretation('Anthropomorphic_Shape', 'Human').
interpretation('Line_Shape', 'Sword').
interpretation('Line_Shape', 'Staff').
interpretation('Triangle_Shape', 'MageCap').
\end{verbatim}
}}

Rules that define how to interpret scenes can be presented, in a user-friendly and simplified form, as \verb|rule(SceneInt, ImgList){Cond}|, stating that the scene consisting of sub-images listed in \verb|ImgList| should be interpreted as \verb|SceneInt| based on conditions \verb|Cond|. The conditions involve the interpretations of sub-images in \verb|ImgList| and the spatial relations between/among them. 

\begin{table}[h]
\begin{flushleft}
\texttt{ interpretationRule ::= {\textbf{ rule(}}sceneInt {\textbf{, [}} imgList {\textbf{] )\{}}cond{\textbf{\}}} }\\
\texttt{ sceneId ::=} {\em uppercase alphanumeric string}\\
\texttt{ imgId ::=} {\em uppercase alphanumeric string}\\
\texttt{ interprId ::=} {\em uppercase alphanumeric string}\\
\texttt{ sceneInt ::= {\textbf{'}} sceneId {\textbf{'}} }\\
\texttt{ imgList ::=  imgId | imgId {\textbf{,}} imgList }\\
\texttt{ constraint ::= interprId{\textbf{(}} imgId {\textbf{)}} | property{\textbf{(}} imgList {\textbf{)}}}\\
\texttt{ disjcond ::=  constraint {\textbf{or}} constraint | constraint {\textbf{or}} disjcond }\\
\texttt{ cond ::=  constraint | {\textbf{(}} disjcond {\textbf{)}} | constraint {\textbf{;}} cond }\\
\texttt{ property ::= {\textbf{ horizontal }} | {\textbf{ vertical }} |  {\textbf{ diagonal }} | {\textbf{ disjoint }} | ...}\\
\end{flushleft}
\caption{User-friendly modelling language for scene interpretation rules: boldface symbols are terminals; alphanumeric uppercase strings are defined in the usual way; properties should include at least the geometric binary relations listed in the BNF, but unary properties such as the image color or source, and n-ary properties such as belonging to the same group, could be added.}
\label{tab:BNF}
\end{table}

Domain experts may use the user-friendly syntax, whose BNF is presented in Table \ref{tab:BNF}, which can be automatically translated into standard Prolog\footnote{The translation has not been implemented so far, but the automatic translation rules are easy to devise, with {\textbf{;}} translated into {\textbf{,}}, {\textbf{or}} translated into {\textbf{;}}, testing of geometric properties translated into \texttt{jpl\_call} with the property to be tested as argument.}.  
% Rules are written in Prolog. 
% A list containing $N$ elements is represented as \texttt{[E1, E2, ... EN]}.
% Lowercase words and words enclosed within single quotation marks represent constants; uppercase words are logical variables: when used during execution, any occurrence of the same variable must be bound to the same constant value. 
%

As an example, the first rule below can be read as ``if token \verb|X| has been interpreted as a human figure, and if token \verb|Y| has been interpreted as a sword, and if  \verb|X| and \verb|Y| are positioned horizontally, then they form a scene representing a \verb|Warrior|''. The second rule is similar, but states when two tokens represent a \verb|Shepherd|. %: note that this is an example where, considering an initial image made of two sub-images, two rules (two different interpretations) can be applied, depending on how one of the sub-images (the one with \verb|'Line_Shape'| classification, not reported for readability) is interpreted (as a \verb|Sword| or as a \verb|Staff|).

% The actual Prolog rules used by OntoScene convey the very same meaning and structure as these ones; they are less readable since they use the concrete Prolog syntax and JPL calls to spatially-related methods based on JTS. In the Prolog rules we abuse Prolog notation by using \verb|ImgInt(X)|  to mean that token \verb|X| has been interpreted as \verb|ImgInt|.

{\small{
\begin{verbatim}
rule('Warrior', [X,Y]) {	               rule('Shepherd', [X,Y]) {
  Human(X);                                    Human(X);
  Sword(Y);                                    Staff(Y);
  horizontal(X,Y);                             horizontal(X,Y);
}                                               }
\end{verbatim}
}}

\begin{figure}[h]
\begin{center}
\includegraphics[width=9cm]{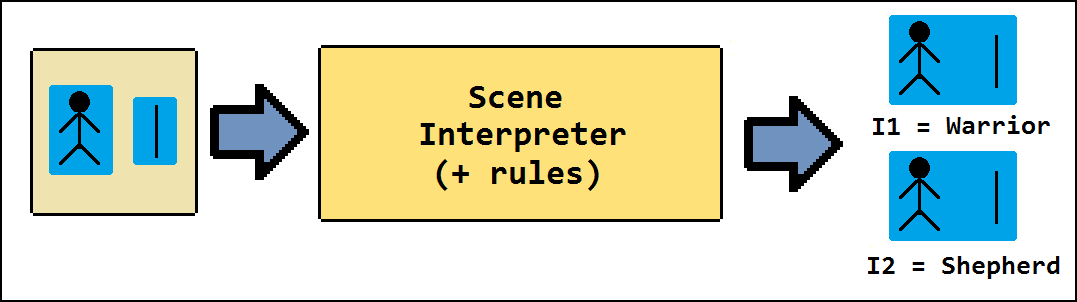}
\caption{Input scene: first example.}
\label{fig:19}
\end{center}
\end{figure}

Let us suppose that the \texttt{Classifier} has classified the left-most sub-image in Figure \ref{fig:19} as an \texttt{An\-thro\-po\-mor\-phic\_Shape} and  the right-most as a \verb|Line_Shape|, and the rules above have been loaded into the \texttt{SceneInterpreter} module. Let us also assume that the horizontal geometric relationship holds between the two sub-images. \texttt{SceneInterpreter} generates two interpretations: \verb|Warrior(I1)| and \verb|Shepherd(I2)|. Interpretation I1 is generated when the right-most sub-image is interpreted as a \verb|Sword| (because of the rule for \verb|Warrior|) while I2 is generated when it is interpreted as a \verb|Staff| (because of the rule for \verb|Shepherd|).

\begin{figure}[h]
\begin{center}
\includegraphics[width=10cm]{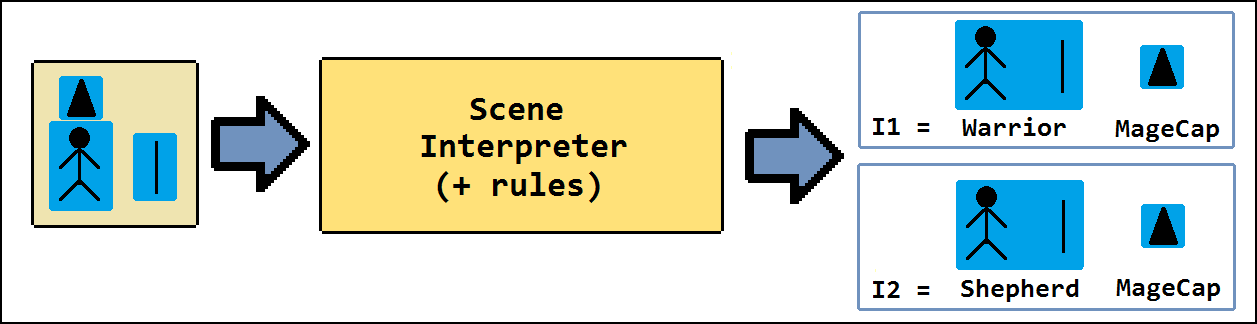}
\caption{Input scene: second example.}
\label{fig:20}
\end{center}
\end{figure}

Figure \ref{fig:20} shows the interpretations of the same scene shown in Figure \ref{fig:19} where a triangular shape has been added on top of the human figure. \texttt{SceneInterpreter} always tries to aggregate as many tokens as possible, but since there are no rules involving the mage cap together with the other elements of the figure, the computed interpretations are those output before, where the triangle is interpreted as a ``stand-alone'' element.  

\begin{figure}[h]
\begin{center}
\includegraphics[width=10cm]{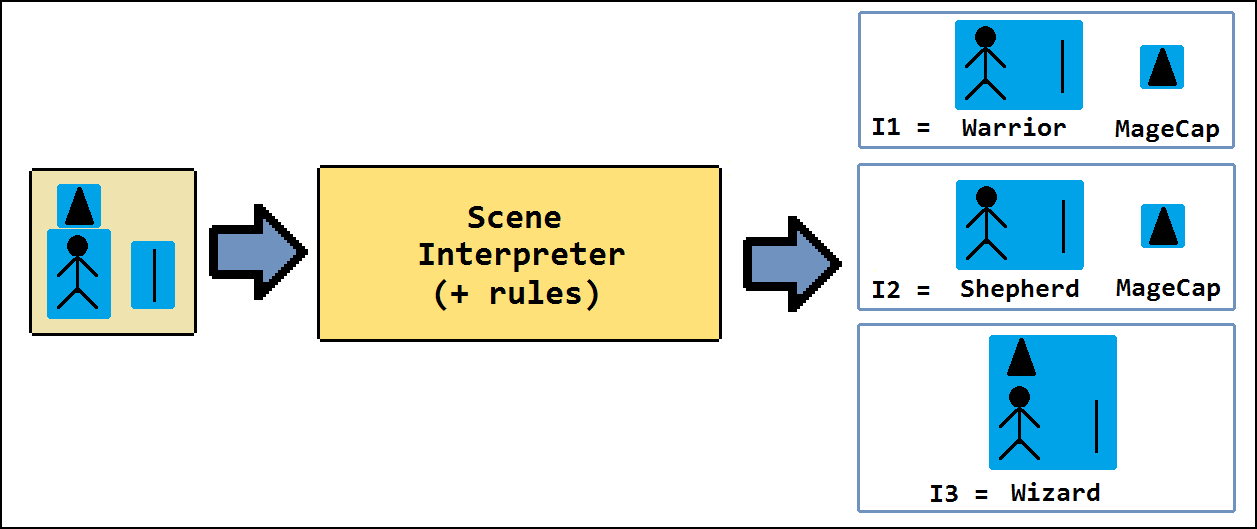}
\caption{Input scene: third example.}
\label{fig:21}
\end{center}
\end{figure}

If another rule were available, 
{\small{
\begin{verbatim}
rule('Wizard', [X,Y,Z]) {
  Human(X);
  MageCap(Y);
  Staff(Z);
  vertical(X,Y);
  horizontal(X,Z);
}
\end{verbatim}
}}
stating that a wizard is a human figure with a magician's hat on top and a stick placed horizontally, then the \texttt{SceneInterpreter} output would be the one shown in Figure \ref{fig:21}. 
%More details on the \texttt{SceneInterpreter} functioning are provided in Section \ref{sec:SceneInterpreter}.
%, where the Prolog implementation of the scene interpretation rules is used. 

%3.2.4 Integrating the Framework into a JADE MAS
\subsection{Making OntoScene Functionalities Available to JADE: the OntoSceneAgent}
\label{sec:3.2.4}

OntoScene has been designed to be a component able to offer the interpretation service, and to be naturally integrated within a JADE MAS.
%(Figure \ref{fig:25}).
%\begin{figure}[h]
%\begin{center}
%\includegraphics[width=7cm]{images/25.png}
%\caption{OntoSceneAgent.}
%\label{fig:25}
%\end{center}
%\end{figure}
%
The steps required to perform the integration in a JADE MAS are: 
\begin{itemize}
\item to integrate the ontology used in the MAS with the OntoScene ontology in order to allow all agents to be aware of the input and output concepts used within the framework and allow their exchange via JADE messages; 
\item to add a new JADE action representing the interpretation of a scene (\verb|InterpretAction|): we achieved both these two steps thanks to the OBJ5.0 framework \cite{WOA18};
\item to implement an agent acting as an interface between the other agents and OntoScene; this agent (the OntoSceneAgent) waits for an agent $A$ to send a request to perform the action \verb|InterpretAction|, with an input scene, calls the \texttt{SceneInterpreter} module on it, and returns the scene interpretations to $A$. %We tested this agent with the IndianaMas system \cite{briola2017agent}.

\end{itemize}

Since this issue is not central to the paper, which focuses on the implementation of the OntoScene framework, we do not expand it further.
%in its body: for the interested readers, technical details are provided in Appendix A.

% 3.2.3 The ontology of the framework
\subsection{The OntoScene Ontology}
\label{sec:3.2.3}

\begin{figure}[h]
\begin{center}
\includegraphics[width=7cm]{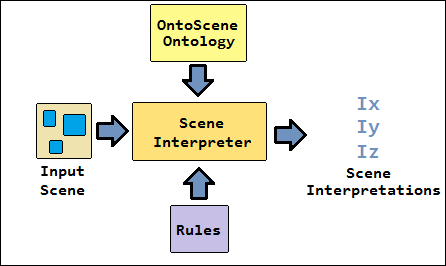}
\caption{The OntoScene Ontology in the OntoScene context.}
\label{fig:22}
\end{center}
\end{figure}

To formalize the OntoScene domain and make interoperability among the many modules involved in the framework possible (Figure \ref{fig:22}), an ontology called OntoScene Ontology has been designed and implemented. 

The OntoScene Ontology is aimed at ensuring modularity and domain-independence: the user can extend it by adding more  domain concepts from existing or new ontologies. In fact, concepts such as \verb|Classification| and \verb|Interpretation|, which characterize the ontology (see Section \ref{sec:ontosceneOnto} for more details) are necessarily domain-specific: by changing the domain ontology that extends the OntoScene Ontology, and consistently changing the interpretation rules, the user can modify the application domain while leaving the OntoScene core functionalities unchanged.

\subsection{Back to the Initial Scenario}

\begin{figure}[h]
\begin{center}
\includegraphics[width=4cm]{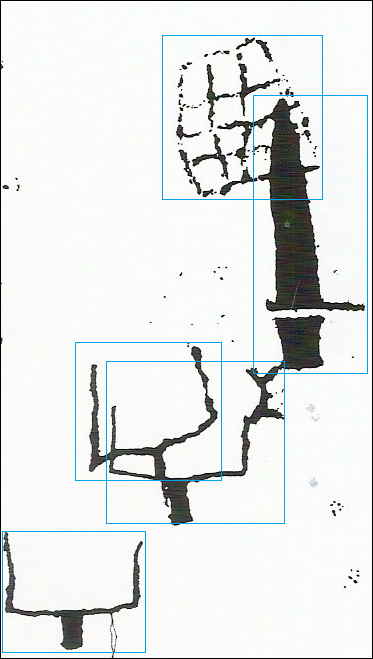} 
\caption{The Scene from the book by de Lumley and Echassoux   ({\color{blue}{de Lumley and Echassoux 2009}}) with detected bounding boxes.}
\label{fig:complexscene}
\end{center}
\end{figure}
Thanks to the components mentioned in the previous sections, we can obtain the bounding boxes shown in Figure \ref{fig:complexscene} and the interpretations, represented in a way that should be intuitive enough and that will be explained in details in Section \ref{sec:SceneInterpreter}, below:
\begin{verbatim}
I1 = [Storm_God(Reticulum-0, Dagger-1),
Group_Of_Corniforms(Corniform-2,Corniform-3,Corniform-4)].

I2 = [Storm_God(Reticulum-0, Dagger-1),
Bull_God(Corniform-2,Corniform-3),Corniform-4].
\end{verbatim}
Given that \verb|bb(X,Y,W,H)| states the \verb|X| and \verb|Y| coordinates of the top- and left-most corner of the bounding box plus its \verb|W|idth and \verb|H|eigh, this is the actual result we get by running OntoScene on the input
\begin{verbatim}
image(0, bb(161, 12, 165, 167), [class('Reticulum_Class', 1.0)]).
image(1, bb(257, 68, 109, 281), [class('Dagger_Class', 1.0)]   ).
image(2, bb(86, 323, 162, 129), [class('Up_Corn_Class', 1.0)]  ).
image(3, bb(107, 337, 181, 162),[class('Up_Corn_Class', 1.0)]  ).
image(4, bb(3, 506, 144, 23),   [class('Up_Corn_Class', 1.0)]  ).
\end{verbatim}
and includes the correct interpretation \texttt{I1} provided by Henry de Lumley and Annie Echassoux, two archaeologists who spent their life on rock art interpretation, in the book from which the image is taken. 
%Designing and implementing complex interpretation rules would allow us to refine the output and discard \texttt{I2}, but the results we get even with very simple rules are promising, as discussed in Section \ref{sec:CaseStudy}.

\section{Modelling and Implementing Domain and Spatial Knowledge}
\label{sec:ontoandgeom}

\subsection{Domain Knowledge}
\label{sec:ontosceneOnto}

\begin{figure}[h]
\begin{center}
\includegraphics[width=6cm]{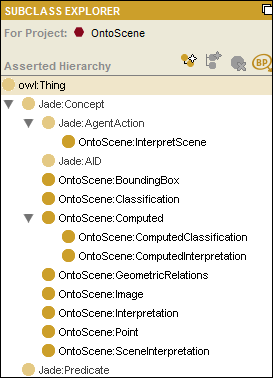}
\caption{The OntoScene ontology.}
\label{fig:46}
\end{center}
\end{figure}

The OntoScene ontology imports the JADE template ontology, needed to let the ontology be directly usable by JADE, as described in Section \ref{sec:background}. It contains all the concepts that \texttt{Sce\-ne\-In\-ter\-pre\-ter} uses during a scene interpretation, and is designed to be extended with an existing domain ontology to integrate \texttt{SceneInterpreter} within a MAS in a transparent way. The classes provided by the OntoScene ontology are shown in Figure \ref{fig:46}.

\paragraph{Point.}
%
%\begin{figure}[h]
%\begin{center}
%\includegraphics[width=4cm]{images/47.png}
%\caption{The Point class.}
%\label{fig:47}
%\end{center}
%\end{figure}
%
The \texttt{Point} class contains two single {\em float} properties \texttt{X} and \texttt{Y}.
%as shown in Figure \ref{fig:47}.

\paragraph{BoundingBox.} 
%
%\begin{figure}[h]
%\begin{center}
%\includegraphics[width=7cm]{images/48.png}
%\caption{The BoundingBox class.}
%\label{fig:48}
%\end{center}
%\end{figure}
%
The \texttt{BoundingBox} class, abbreviated as \texttt{BB}, represents the rectangle that bounds a single image.
%It contains four single properties (\texttt{Point}) representing the four vertices of the \texttt{BB} (\texttt{TopLeftPoint}, \texttt{TopRightPoint}, \texttt{BottomRightPoint} and \texttt{BottomLeftPoint}) and two {\em float} properties to represent \texttt{Width} and \texttt{Height}. %, as shown in Figure \ref{fig:48}.

\paragraph{ComputedClassification.}
%
%\begin{figure}[h]
%\begin{center}
%\includegraphics[width=8.5cm]{images/49.png}
%\caption{The ComputedClassification class.}
%\label{fig:49}
%\end{center}
%\end{figure}
%
The \texttt{ComputedClassification} class represents a classification computed  by the \texttt{Classifier} along with its confidence. It contains the single properties \texttt{identi\-fied\-Clas\-sification} with range \texttt{Classification} and \texttt{confidence} with range {\em float}. 
% as shown in Figure \ref{fig:49}.

\paragraph{ComputedInterpretation.}
%
%\begin{figure}[h]
%\begin{center}
%\includegraphics[width=8.5cm]{images/50.png}
%\caption{The ComputedInterpretation class.}
%\label{fig:50}
%\end{center}
%\end{figure}
%
The \texttt{ComputedInterpretation} class represents an interpretation computed by \texttt{SceneInterpreter} with the associated confidence and its size, namely how many input images have been aggregated. It contains the single properties \texttt{iden\-ti\-fied\-In\-ter\-pre\-tation}  with range \texttt{Interpretation}, \texttt{confidence} with range {\em float} and \texttt{size}, with range {\em int}. % (Figure \ref{fig:50}). 

\paragraph{Classification and Interpretation.}
\texttt{Classification} and \texttt{Interpretation} are two classes without any property and their meaning is the intuitive one. To allow \texttt{SceneInterpreter} to interpret an input scene, some classes from the domain ontology must necessarily extend these two classes with domain-specific classifications and interpretations.

\paragraph{GR.}
The \texttt{GR} class is used as a container for methods representing geometric relationships, to be called within the body of rules through predicates offered by the JPL library.
\texttt{SceneInterpre\-ter} uses an internal class called \texttt{GeometricRe\-la\-tions\-Impl} with the implementation of those methods that we used to test the program. More sophisticated implementations can be used instead of the ones we provide: the Java Method Mapper panel of OBG5.0 allows the developer to create methods under the \texttt{GR} class and export their interface, in order to be implemented.

\paragraph{Image.}
\begin{figure}[h]
\begin{center}
\includegraphics[width=8.5cm]{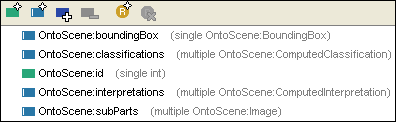}
\caption{The Image class. Multiple stands for List of.}
\label{fig:51}
\end{center}
\end{figure}
The \texttt{Image} class represents a basic or composite scene.
It contains a single \texttt{id} property of type {\em int} that acts as an identifier, a single \texttt{boundingBox} property of type \texttt{BoundingBox} for the \texttt{BB}, a multiple \texttt{classification} property of type \texttt{ComputedClassification} listing all the  classifications assigned by the \texttt{Classifier} to the image in the scene, a multiple \texttt{interpretation} property of type \texttt{ComputedInterpretation} including the interpretations computed by \texttt{SceneInterpreter} and a multiple \texttt{subParts} property of type \texttt{Image} that contains all the sub-images that form the image, as shown in Figure \ref{fig:51}.

The \texttt{Image} class is the main data structure used by \texttt{SceneInterpreter} to keep track of the relationship between Prolog scenes represented as Prolog facts, and Java scenes represented as instances of the Java \texttt{Image} class. Each time a new node (namely, a new scene) is added to the scene graph, the corresponding \texttt{Image} instance is also created inside it: there is a one-to-one association between each node in the scene graph and an \texttt{Image} instance.
In the sequel we will usually use {\em image} and {\em sub-image} when we refer to data representations on the Java side, and {\em scene} and {\em sub-scene} when we refer to the Prolog side. 

In order to work properly, \texttt{SceneInterpreter} expects input images with these features:
\begin{itemize}
\item \texttt{id} and \texttt{boundingBox} fields instantiated;
\item \texttt{classifications} instantiated with a list of one or more classifications;
\item empty \texttt{interpretations} list;
\item empty \texttt{subParts} list.
\end{itemize}

The association between classifications and interpretations is computed by the Prolog engine via the  \texttt{interpretation/2} predicate
% , which provides the actual implementation of the \texttt{interpreted\_as(X,Y)} predicate 
introduced in Section \ref{sec:3.2.2}.

After the creation, via the aggregation rules, of a composite scene in Prolog, \texttt{Sce\-ne\-In\-ter\-pre\-ter} creates a new \texttt{Image} object that corresponds to the new scene and has these features:
\begin{itemize}
\item \texttt{id} field instantiated with a new unique identifier;
\item \texttt{boundingBox} obtained by merging the \texttt{BB}s of the sub-scenes;
\item empty \texttt{classifications} list, as only basic scenes have a classification;
\item \texttt{interpretations} list containing the computed interpretations;
\item \texttt{subParts} instantiated with the list of the sub-images.
\end{itemize}

\paragraph{SceneInterpretation.}
%
%\begin{figure}[h]
%\begin{center}
%\includegraphics[width=7cm]{images/52.png}
%\caption{The SceneInterpretation class.}
%\label{fig:52}
%\end{center}
%\end{figure}
%
The \texttt{SceneInterpretation} class 
%(Figure \ref{fig:52}) 
represents an interpretation of the input scene. It contains a \texttt{composedBy} property of type \texttt{Image} that contains all the images of the interpretation in the format presented above, corresponding to the scenes that can coexist.

The agent that, upon reception of an \texttt{InterpretScene} action presented below, is required to provide a scene interpretation, returns a \texttt{SceneInterpretation} list.
%, as the possible interpretations are usually more than one.

\paragraph{InterpretScene Action.}
%
%\begin{figure}[h]
%\begin{center}
%\includegraphics[width=7cm]{images/53.png}
%\caption{The InterpretScene AgentAction class.}
%\label{fig:53}
%\end{center}
%\end{figure}
%
The \texttt{InterpretScene} class extends the JADE \texttt{AgentAction} class and represents the action of requesting the interpretation of an input scene.
It contains a multiple property \texttt{inputImages} of type  \texttt{Image} representing the images in the input scene and two {\em boolean} properties, \texttt{distinct} and \texttt{filtered}, which refer to the interpretation mode. When \texttt{distinct} mode is selected, all the scenes in the final  list of \texttt{SceneInterpretation} must be distinct Java objects, in order to obtain a readable and writable data structure. When \texttt{filtered} mode is on, only filtered interpretations are returned.
      
% \paragraph{OntoScene inside JADE.}
% Figure \ref{fig:54} shows a JADE MAS which integrates the OntoScene framework.
% %
% \begin{figure}[h]
% \begin{center}
% \includegraphics[width=7.5cm]{images/54.png}
% \caption{A JADE MAS.}
% \label{fig:54}
% \end{center}
% \end{figure}
% %
% Agent \texttt{X} sends an ACL message containing the \texttt{InterpretScene} action to Agent \texttt{Y}, in charge for scenes interpretation.
% Agent \texttt{Y} computes the \texttt{SceneInterpretation} list by calling the SceneInterpreter API methods and returns the result.

\begin{figure}[!htb]
   \begin{minipage}{0.48\textwidth}
     \centering
     \includegraphics[width=.7\linewidth]{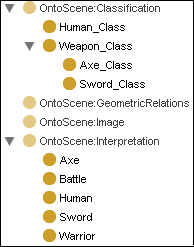}
     \caption{Domain-dependent concepts that extend \texttt{OntoScene} classes.}
     \label{fig:56}
   \end{minipage}\hfill
   \begin{minipage}{0.48\textwidth}
     \centering
     \includegraphics[width=.4\linewidth]{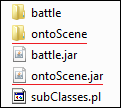}
     \caption{The files generated by OBG5.0.}
     \label{fig:57}
   \end{minipage}
\end{figure}

\paragraph{An example Ontology: Battle.}
The \texttt{Bat\-tle} ontology models a simplified domain that will be used in the next section. 
%After opening the OntoScene.owl ontology template (which in turn requires uploading the JADE template ontology) in Protégé, a new Battle Ontology was created (with the 'Create empty imported ontology ...' tab of the tab Metadata) first empty and selected as the main ontology.
%
%Figure \ref{fig:55} shows the selection (with the pencil icon) of the Battle Ontology.
%
%\begin{figure}[h]
%\begin{center}
%\includegraphics[width=7cm]{images/55.png}
%\caption{BOOO.}
%\label{fig:55}
%\end{center}
%\end{figure}
%
%\texttt{Battle} must be created as a new ontology in Prot\'eg\'e and imported by \texttt{OntoScene.owl}.
%
%\begin{figure}[h]
%\begin{center}
%\includegraphics[width=4.5cm]{images/56.png}
%\caption{Domain-dependent concepts which extend \texttt{OntoScene} classes.}
%\label{fig:56}
%\end{center}
%\end{figure}
%
Figure \ref{fig:56} shows how the \texttt{Classification} and \texttt{Interpretation} classes of \texttt{OntoScene} can be sub-classed by classes characterizing the Battle domain, where armed warriors fight using swords or axes.
%
%Note:
%As for the use of SceneInterpreter, it is required to extend at least the Classification and Interpretation classes with domain classes.
%
%\begin{figure}[h]
%\begin{center}
%\includegraphics[width=2.8cm]{images/57.png}
%\caption{The files generated by OBG5.0.}
%\label{fig:57}
%\end{center}
%\end{figure}
%
The Java files generated by OBG5.0 are shown in Figure \ref{fig:57}.
%: The \texttt{ontoScene} folder contains all the Java interfaces and concrete classes generated by OBG5.0. 
%\texttt{SceneInterpreter} uses the interfaces contained in \texttt{ontoScene.jar} by using the Java \texttt{import} directive, like \texttt{import ontoScene.OntoScene\_Image}. 
%assuming that the resulting package will be named \texttt{battle.jar}. In addition to the \texttt{battle}, \texttt{battle.jar}, and \texttt{subClasses.pl} files (see Section \ref{sec:background}), the \texttt{ontoScene} folder and the \texttt{ontoScene.jar} packages are generated.

%Note:
%The purpose of splitting the ontology classes into two separate jar files is to allow the user to choose (or maintain) any name for the domain ontology and its jar (battle.jar in this example). If just the name of the package is chosen 'onScene', OBG5.0 does not split jar files and behaves normally (creating a single ontoScene.jar file).

\subsection{Spatial Knowledge}
\label{subsec:geomrel}

To interpret scenes with \texttt{SceneInterpreter}, the user must identify the required geometric relationships and must create methods in the \texttt{GR} class of the OntoScene ontology to represent them. 
%Methods may have an arbitrary signature, but the Prolog interpretation rules must be consistent with them. Once we get the \texttt{GR} interface and the \texttt{DefaultGR} empty implementation from OBG5.0, it is necessary to implement a class that defines the actual body of the methods. 
If the user has no special requirements, (s)he can use the \texttt{GRImpl} we provide with the framework. 
Implementing geometric relationships is not easy, because different domains may need different relationships.  An exception are topological relationships (disjoint, overlap, etc.) for which known mathematical formalisms exist. We used the JTS library to implement the following ones:
\begin{small}
\begin{verbatim}
Boolean horizontal (BB bb1, BB bb2, String pos)
Boolean horizontal (BB bb1, BB bb2)
Boolean vertical (BB bb1, BB bb2, String pos)
Boolean vertical (BB bb1, BB bb2)
Boolean diagonal (BB bb1, BB bb2, String pos)
Boolean diagonal (BB bb1, BB bb2)
Boolean disjoint (BB bb1, BB bb2)
Boolean overlap (BB bb1, BB bb2)
Boolean contains (BB bb1, BB bb2)
Boolean absNear (BB bb1, BB bb2, float th) /* absolute proximity */
Boolean relNear (BB bb1, BB bb2, float th) /* relative proximity */
Boolean absGroup (List <BB> bbs, float th) /* group, using absNear */
Boolean relGroup (List <BB> bbs, float th) /* group, using relNear */
\end{verbatim}
\end{small}

\paragraph{Horizontal, vertical and diagonal relationships.}
The parameters of these methods are two \texttt{BB}s and -- optionally -- a string indicating the position that \texttt{bb1} must have w.r.t. \texttt{bb2}. The position may be {\em right} or {\em left} for horizontal, {\em up} or {\em down} for vertical and {\em se}, {\em sw}, {\em ne}, {\em nw} for diagonal.
For example, \texttt{diagonal(bbx, bby, ne)} is \texttt{true} if \texttt{bbx} is positioned north-east w.r.t. \texttt{bby}.

\paragraph{Topological relationships disjoint, overlap and contain.}
These methods take two \texttt{BB}s \texttt{bbx} and \texttt{bby} in input and answer whether \texttt{bbx} \texttt{rel} \texttt{bby} holds. For example, \texttt{contains(bbx,bby)} is \texttt{true} if \texttt{bbx} contains \texttt{bby}.

\paragraph{Absolute proximity AbsNear and relative proximity RelNear.} Besides the two \texttt{BB}s, these methods also have a third parameter to state the threshold under which the two \texttt{BB}s are considered ``close''. This threshold therefore defines the proximity semantics. 

In \texttt{absNear} the threshold indicates an absolute value expressed in an arbitrary measure unit determined by the domain expert such as pixels, centimeters, etc. For example, assuming pixels as the measure unit, \texttt{absNear(bbx, bby, 10.0)} is \texttt{true} if the absolute distance between the edges of \texttt{bbx} and \texttt{bby} is less than 10px.

\begin{figure}[h]
\begin{center}
\includegraphics[width=4.5cm]{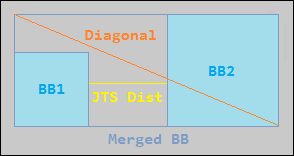}
\caption{Graphical representation of the formula used by \texttt{relNear}.}
\label{fig:130} 
\end{center}
\end{figure}
% %
In \texttt{relNear} the threshold indicates a relative value between 0 and 1.0. This allows us to define  ``proximity'' in a way robust to the image scaling. For example, \texttt{relNear(bbx, bby, 0.2)} is true if $X \leq 0.2$, where $X$ is the value of some expression that the user can define. The one we implemented is explained in Figure \ref{fig:130}: we compute \texttt{JTSDist}, namely the distance between \texttt{bbx} and \texttt{bby} computed by JTS, we merge \texttt{bbx} and \texttt{bby} into \texttt{mbb}, we compute \texttt{Diagonal}, namely the length of  \texttt{mbb} diagonal. $X$ is  \texttt{JTSDist/Diagonal}. If both bounding boxes are scaled by a factor \texttt{F}, \texttt{relNear(bbx*F, bby*F, 0.2)} is the same as \texttt{relNear(bbx, bby, 0.2)}, making the definition invariant w.r.t. scaling.

Finally, the \texttt{absGroup} and \texttt{relGroup} methods compute the ``neighborhood'' relationship on a list of \texttt{BB}s  using \texttt{absNear} and \texttt{relNear} respectively.

%\noindent The details of their specification and implementation is provided in Section \ref{appendix2} of the Appendix.

\section{SceneInterpreter}
\label{sec:SceneInterpreter}

The \texttt{Detector} and \texttt{Classifier} modules work on raw images and produce an ``input image'' consisting of bounding boxes associated with possibly many classifications of their content, drawn from an ontology, along with a confidence on that classification. \texttt{SceneInterpreter} takes this classified ``input image'' as input and transforms it into a set of ``basic scenes'', namely triples consisting of (image, classification, interpretation).
% This section describes how the \texttt{SceneInterpreter} module works and provides examples.

% \subsection{The SceneInterpreter Algorithm}

For each input image, \texttt{Sce\-ne\-In\-ter\-pre\-ter} creates as many basic scenes from the (classification, interpretation) pairs as it can. For example, if a sub-image \texttt{Img1} has been classified by the \texttt{Classifier} module as \texttt{C1} or \texttt{C2}, and \texttt{C1} has \texttt{I11} and \texttt{I12} as possible interpretations, while \texttt{C2} can only be interpreted as the \texttt{I21}, three basic scenes are generated:
\begin{verbatim}
basic_scene(Img1, C1, I11).
basic_scene(Img1, C1, I12).
basic_scene(Img1, C2, I21).
\end{verbatim}

%A basic scene represents the interpretation \texttt{IZ} of an image \texttt{ImgX}, assuming that  \texttt{CY} is the correct classification for \texttt{ImgX}. 
%SceneInterpreter works at baseline scenes. All the interpretations found for the same image coexist as base scenes in the initial state of computation.

% The BB identified by the \texttt{Detector} contains the coordinates of the topmost and leftmost vertex (which define its global position within the input scene), the width and the height (which define the container size).
% Different geometric relationships between images can be identified, based on their BBs. For example, two images could be placed horizontally, vertically or diagonally (space relations),  partially overlapping or totally disjointed (topological relationships). This holds for basic scenes as well, since each of them refers to the BB of the associated image.

The scene interpretation rules that drive \texttt{SceneInterpreter} define how to aggregate the elements in a scene, be they atomic sub-images or scenes, depending on the geometric relationships holding among them. We name them  {\em aggregation rules} in the remainder. Aggregation rules have been also called ``scene interpretation rules'' in the paper; in this section we prefer to use ``aggregation'' to clearly differentiate them from the \texttt{interpretation} predicate that will be presented in Section  \ref{serializing}, which associates an interpretation to a basic image, based on its classification.
A composite scene is a scene created by the aggregation of other scenes, which may be in turn basic or composite ones. We talk about scene, without further distinction, when it is not necessary to distinguish whether the scene is a basic or a composite one. 
\texttt{SceneInterpreter} generates a scene graph representing all the scenes that can be derived by applying the aggregation rules to the basic scenes generated from an input image.

%\begin{figure}[h]
%\begin{center}
%\includegraphics[width=5cm]{images/44-bis.png}
%\caption{A scene graph created from an input image with five sub-images.}
%\label{fig:44}
%\end{center}
%\end{figure}

As an example, the figure in Table \ref{tab:graph1}, left,  shows a scene graph resulting from an input scene containing five different sub-images: they have been transformed into five basic scenes (BS1, BS2, BS3, BS4 and BS5), and then into [composite] scenes thanks to the available aggregation rules. For example in this case, by applying some aggregation rule, BS1 and BS2 can be aggregated into CS1. BS2, BS3 and BS4 can be aggregated into CS2, and so on. We point out that BS2 was used by an aggregation rule to form CS1, and by another to form CS2. In the same way, BS4 can be used to form both CS2 and CS3. BS2 and BS4 are called shared scenes.
The scenes graph is oriented (from top to bottom) and acyclic. A top node, or top scene, is a node with no incoming edges.  In the figure in Table \ref{tab:graph1}, left, CS1 and CS4 are top nodes.

\begin{table}[h]
%\begin{scriptsize}
\begin{tabular}{m{6cm}m{7cm}}
\toprule
Input Image & Interpretations\\
\midrule
\includegraphics[width=5cm]{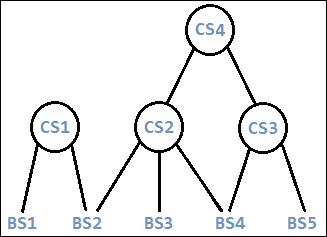} & 

\begin{verbatim}
I1 = [BS1, BS2, BS3, BS4, BS5]
I2 = [CS1, BS3, BS4, BS5]
I3 = [CS1, BS3, CS3]
I4 = [BS1, CS2, BS5]
I5 = [BS1, CS4]
\end{verbatim}
\\
\bottomrule
\end{tabular}
\caption{A scene graph created from an input image with five sub-images, plus the generated interpretations.}
\label{tab:graph1}
%\end{scriptsize}
\end{table}

% \paragraph{Implementation.}

\texttt{SceneInterpreter} core functionalities have been implemented in Prolog.
% implemented using Java and Prolog. Rather than redefining an ad hoc grammar for aggregation rules, we allow the user to write them directly in Prolog. Thanks to backtracking, Prolog is very suitable to implement {\em generate and test} approaches that can be used in the generation of all the correct interpretations, after having applied the aggregation rules. For this reason, Prolog is used by the \texttt{SceneInterpreter} engine, which is implemented in the \texttt{scene\_interpreter.pl} Prolog file. 
%
For efficiency issues, however, geometric relationships have been implemented in Java and are called by Prolog through the JPL library introduced in Section \ref{sec:background}. 
%
%\begin{figure}[h]
%\begin{center}
%\includegraphics[width=7cm]{images/45.png}
%\caption{SceneInterpreter workflow diagram.}
%\label{fig:45}
%\end{center}
%\end{figure}

The steps to be performed to set up \texttt{SceneInterpreter} and to interpret an input image are the following:%, shown in Figure \ref{fig:45}, 
\begin{enumerate}
\item define the aggregation rules in Prolog (done only once);
\item initialize the Java \texttt{SceneInterpreter} module;
\item select the aggregation rules;
\item load a scene composed of a list of images plus their classification (the output of the \texttt{Detector} and \texttt{Classifier} modules), serializing them into basic scenes;
\item apply aggregation rules to create composite scenes and generate the scene graph;
\item generate all the interpretations by calling the \texttt{knuth\_algo\_x} predicate on the scene graph; 
\item filter out interpretations that can be derived from others (optional) and provide the final, sorted result.
\end{enumerate}

The steps from 4 to 7 are discussed in Sections from \ref{serializing} to \ref{filtering-sorting-returning}, respectively. 

\subsection{Serializing Images in Basic Scenes}
\label{serializing}

To allow \texttt{SceneInterpreter} to serialize input images into Prolog scenes, associations between classifications and domain interpretations created under the \texttt{Classification} and \texttt{In\-ter\-pre\-tation} ontology classes must be provided.
The predicate that OntoScene offers to this aim is 
\begin{verbatim}
interpretation/2
interpretation(Class,Inter).
\end{verbatim}
whose meaning is that a picture classified as \texttt{Class} can be interpreted as \texttt{Inter}.
For the classification and interpretation within the Battle domain we defined the following facts:
\begin{verbatim}
interpretation('Human_Class', 'Human').
interpretation('Sword_Class', 'Sword').
interpretation('Axe_Class', 'Axe').
\end{verbatim}
The \texttt{Human\_Class} classification can be directly interpreted as \texttt{Human}, the \texttt{Sword\_Class} as \texttt{Sword} and the \texttt{Axe\_Class} as \texttt{Axe}. During the image serialization, these facts are used by \texttt{SceneInterpreter} to create the basic scenes.% (the scene nodes of the scene graph).
\\

\noindent A predicate called \texttt{scenes/6} is used to represent basic and composite scenes in Prolog.
%\paragraph{The scenes predicate.}
The signature of the predicate is the following:
\begin{verbatim}
scenes/6.
scenes(ID, BB, Class, Inter, Conf, SS).
\end{verbatim}
\begin{itemize}
\item \texttt{ID} is the identifier that Prolog uses to identify scenes\footnote{The \texttt{Id} property of the \texttt{Image} class is used by Java and may be different from \texttt{ID}.};
\item \texttt{BB} is the reference to the Java object representing the \texttt{BoundingBox} of the image in the input scene; %For each new composite scene the merging of its BB is made.
\item \texttt{Class} is the classification of the image from which this scene comes from. The field is instantiated in basic scenes and is empty in composite scenes;
\item \texttt{Inter} is the interpretation of the scene. For basic scenes the variable is instantiated by calling the \texttt{interpretation/2} predicate, while for composite scenes the value to associate with the variable is computed by applying the aggregation rules;
\item \texttt{Conf} is the confidence of the interpretation associated with the scene. For basic scenes whose confidence in the classification is \texttt{C}, \texttt{Conf} is computed as  \texttt{C*(1.0/Count)}, where \texttt{Count} is the number of interpretations associated with the scene. For composite scenes, \texttt{Conf = (Conf1 + Conf2 + ... ConfN) / N} where \texttt{N} is the number of aggregated scenes, and \texttt{ConfX} is the confidence of \texttt{X} scene;
\item \texttt{SS} stands for  \texttt{SubScenes} and is the  list of the \texttt{ID}s of the basic scenes belonging to the scene.
\end{itemize}
%%
%\begin{figure}[h]
%\begin{center}
%\includegraphics[width=6.5cm]{images/59.png}
%\caption{Input scene: example 1.}
%\label{fig:59}
%\end{center}
%\end{figure}
%%

\noindent The serialization algorithm is, in pseudocode, the following:
\begin{verbatim}
InputScene S;
For (Image img: S.getImages ())
  For (Classification class: img.getClassifications ())
    For (Interpretation inter: interpretation (class, inter))
      Assert (scenes (ID, BB, class, inter, Conf, SS))
\end{verbatim}
That is, given an input scene \texttt{S}, for each sub-image \texttt{img} belonging to \texttt{S}, for each classification \texttt{class} of \texttt{img}, for each interpretation \texttt{inter} found by calling the Prolog \texttt{interpretation/2} predicate, the fact \texttt{scene} with suitable arguments is asserted in the Prolog knowledge base, for efficient retrieval. Each individual input image is subdivided into as many basic scenes as the found \texttt{(class, inter)} pairs.

For example, let us suppose that the input scene consists of three sub-images shown in Table \ref{tab:ex59}, classified as \texttt{Human\_Class}, \texttt{Sword\_Class}, and \texttt{Axe\_Class} with maximum confidence.
Images are serialized in three \texttt{scene} Prolog facts as shown in the right part of the table.
%\begin{verbatim}
%scene(0, BB1, 'Human_Class', 'Human', 1.0, [0]).
%scene(1, BB2, 'Sword_Class', 'Sword', 1.0, [1]).
%scene(2, BB3, 'Axe_Class', 'Axe', 1.0, [2]).
%\end{verbatim}

\begin{table}[h]
\begin{scriptsize}
\begin{tabular}{m{6cm}m{7cm}}
\toprule
Input Image & Interpretations\\
\midrule
\includegraphics[width=4.5cm]{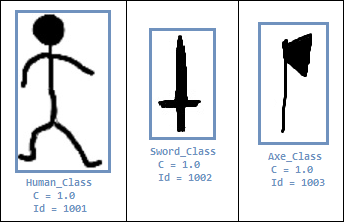} & 

\begin{verbatim}
scene(0, BB1, 'Human_Class', 'Human', 1.0, [0]).
scene(1, BB2, 'Sword_Class', 'Sword', 1.0, [1]).
scene(2, BB3, 'Axe_Class', 'Axe', 1.0, [2]).
\end{verbatim}
\\
\bottomrule
\end{tabular}
\caption{Input Scene, example 1.}
\label{tab:ex59}
\end{scriptsize}
\end{table}

%As can be seen, the OntoScene Ids of the images have been mapped into logical Ids starting with 0. The classifications and interpretations were associated with the predicate interpretation / 2 and the confidence with the formula (1.0 * (1.0 / 1). 
%    
In the first example, each classification is associated with only one interpretation defined by the domain ontology, but in general there could be a one-to-many relationship. 
\begin{figure}[h]
\begin{center}
\includegraphics[width=7cm]{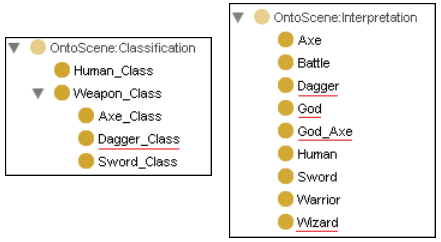}
\caption{Extending the Battle ontology with new classifications and interpretations.}
\label{fig:60}
\end{center}
\end{figure}
Let us now make the example more complex by adding the \texttt{Dagger\_Class} classification and the \texttt{Dagger}, \texttt{God}, \texttt{God\_Axe}, and \texttt{Wizard} interpretations (Figure \ref{fig:60}).
New \texttt{interpretation} facts could be defined as:
\begin{verbatim}
interpretation('Human_Class', 'God').
interpretation('Human_Class', 'Wizard').
interpretation('Axe_Class', 'God_Axe').
interpretation('Dagger_Class', 'Dagger').
\end{verbatim}
%%
%\begin{figure}[h]
%\begin{center}
%\includegraphics[width=7cm]{images/61.png}
%\caption{Input scene: example 2.}
%\label{fig:61}
%\end{center}
%\end{figure}
%%

%\begin{verbatim}
%scene(0, BB1, 'Human_Class', 'Human', 0.33, [0]).
%scene(0, BB1, 'Human_Class', 'God', 0.33, [0]).
%scene(0, BB1, 'Human_Class', 'Wizard', 0.33, [0]).
%scene(1, BB2, 'Sword_Class', 'Sword', 0.8, [1]).
%scene(1, BB2, 'Dagger_Class', 'Dagger', 0.5, [1]).
%scene(2, BB3, 'Axe_Class', 'Axe', 0.5, [2]).
%scene(2, BB3, 'Axe_Class', 'God_Axe', 0.5, [2]).
%\end{verbatim}

\begin{table}[h]
\begin{scriptsize}
\begin{tabular}{m{6cm}m{7cm}}
\toprule
Input Image & Interpretations\\
\midrule
\includegraphics[width=4.5cm]{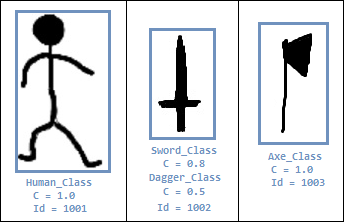} & 

\begin{verbatim}
scene(0, BB1, 'Human_Class', 'Human', 0.33, [0]).
scene(0, BB1, 'Human_Class', 'God', 0.33, [0]).
scene(0, BB1, 'Human_Class', 'Wizard', 0.33, [0]).
scene(1, BB2, 'Sword_Class', 'Sword', 0.8, [1]).
scene(1, BB2, 'Dagger_Class', 'Dagger', 0.5, [1]).
scene(2, BB3, 'Axe_Class', 'Axe', 0.5, [2]).
scene(2, BB3, 'Axe_Class', 'God_Axe', 0.5, [2]).
\end{verbatim}
\\
\bottomrule
\end{tabular}
\caption{Input Scene, example 2.}
\label{tab:ex61}
\end{scriptsize}
\end{table}

In a second example shown in Table \ref{tab:ex61}, the image in the center can be classified in two ways: \texttt{Sword\_Class} and \texttt{Dagger\_Class} (each having only one interpretation), while %, with its serialization as \texttt{scene} facts is reported on the right.
the image on the left has one classification \texttt{Human\_Class} with three interpretations (\texttt{Human}, \texttt{God} and \texttt{Wizard}). The image on the right has one classification (\texttt{Axe\_Class}) and two interpretations (\texttt{Axe} and \texttt{God\_Axe}). 
The confidence is \texttt{1.0*(1.0/3) = 0.33} for each interpretation of the left image,  %The image in the center has two distinct classifications, each having only one interpretation. 
is \texttt{0.8*(1.0/1)} and \texttt{0.5*(1.0/1)}) for the two interpretations of the image in the center, and is \texttt{1.0*(1.0/2)} for the image on the right.

\subsection{Applying aggregation rules for composite scenes and updating the scene graph}

After defining the \texttt{interpretation/2} predicate for the basic scenes,  it is necessary to create aggregation rules for composite scenes. We use the predicate \texttt{rules/2}, stating which scenes should be aggregated, which geometric relationships between their BBs should hold, and computing a list of \texttt{scene} facts that \texttt{SceneInterpreter} uses to generate (possibly) a new composite scene, with interpretation \texttt{Inter}. 
%For efficiency issues, the portion of the body of this rule that selects the scenes to be aggregated, and the portion which checks geometric relationships are separated by a special \texttt{scene\_ok} predicate whose sole purpose is to tag the end of the first part of the algorithm to the engine.

The clauses for the \texttt{rule} predicate, which are semi-automatically compiled into Prolog from the user-friendly modelling language presented in Table \ref{tab:BNF}, follow this pattern: 
\begin{verbatim}
rule(Inter, Scenes): -
% Part 1: Selects the scenes to be aggregated in the Scenes list
% Part 2: Computes geometric relationships
\end{verbatim}

These rules convey the very same meaning and structure as those presented in Section \ref{sec:3.2.2}; they are less readable since they use the concrete Prolog syntax and JPL calls to spatially-related methods based on JTS. For sake of clarity we will abuse Prolog notation by using \verb|ImgInt(X)|  to mean that token \verb|X| has been interpreted as \verb|ImgInt|.
The (manual) process for compiling the user-friendly modelling language into Prolog is not optimized: this can be noticed for example in the usage of \texttt{append} in Table \ref{tab:WarriorScene2}, which could be avoided by using unification instead. While losing in elegance of the resulting code, the naif manual compilation produced rules which follow the same pattern and gave useful hints on how they implement the automatic compilation, which will be addressed as a close future work. 

\noindent Two utility predicates used inside \texttt{rule} clauses are
\begin{verbatim}
relations/1
relations(GR).
\end{verbatim}
and 
\begin{verbatim}
subclass_of/2
subclass_of(Class, SubClass).
\end{verbatim}

\noindent \texttt{relations(GR)} unifies \texttt{GR} with a reference to the implementation of the interface for the geometric relations, instantiated during the OntoScene configuration stage via a call to \texttt{jpl\_new/3}.
In our code, the assertion of the \texttt{relations(GR)} predicate is achieved via 
\begin{verbatim}
assert_relations :-
  jpl_new('onto_impl.GeometricRelationsImpl', [], GR),
  assert(relations(GR)).
\end{verbatim}
\noindent Other OntoScene users might use our implementation of geometric relations, provided via the \texttt{'onto\_impl.GeometricRelationsImpl'} interface, or develop a new one.

The \texttt{subclass\_of(Class, SubClass)} is a predicate exported with OBG5.0: it allows scenes to be analyzed by exploring hierarchies of classes in the ontology, in particular those below the  \texttt{Clas\-si\-fi\-ca\-tion} and \texttt{In\-ter\-pre\-ta\-tion} classes.

Each scene generated by applying one aggregation rule is asserted as a node of the scene graph which is modelled via the \texttt{image\_graph(G)} fact, and which is updated any time a new scene interpretation is computed for a given image, reaching at the end the structure exemplified in Table \ref{tab:graph1}. 

In the sequel we provide some examples of aggregation (scene interpretation): \texttt{near} is used as an abbreviation for \texttt{absNear} and lengths are expressed in pixels.

\paragraph{Example 1: Warrior Scene (Human + Weapon).}
%\begin{figure}[h]
%\begin{center}
%\includegraphics[width=8cm]{images/63.png}
%\caption{Warrior Scene (Human + Weapon).}
%\label{fig:63}
%\end{center}
%\end{figure}

A generic \texttt{Warrior} scene can be defined as a combination of a \texttt{Human} scene and a basic scene classified as \texttt{X}, where \texttt{X} is a sub-class of \texttt{Weapon\_Class} in the ontology (Table \ref{tab:WarriorScene2}).

%\begin{verbatim}
%rule('Warrior', Scenes) :-
%  scene(ID1, BB1, Class1, 'Human', Conf1, SS1),
%  subclass_of('Weapon_Class', Class),
%  interpretation(Class, Weapon),
%  scene(ID2, BB2, Class2, Weapon, Conf2, SS2),
%  append([scene(ID1, BB1, Class1, 'Human', Conf1, SS1)],
%         [scene(ID2, BB2, Class2, Weapon, Conf2, SS2)],
%         Scenes),
%  relations(GR),
%  jpl_call(GR, horizontal, [BB1, BB2], @(true)),
%  jpl_call(GR, near, [BB1, BB2, 2.0], @(true)).
%\end{verbatim}

\begin{table}[h]
\begin{scriptsize}
\begin{tabular}{m{5cm}m{8cm}}
\toprule
Example Image & Scene Interpretation Rule\\
\midrule
\includegraphics[width=5cm]{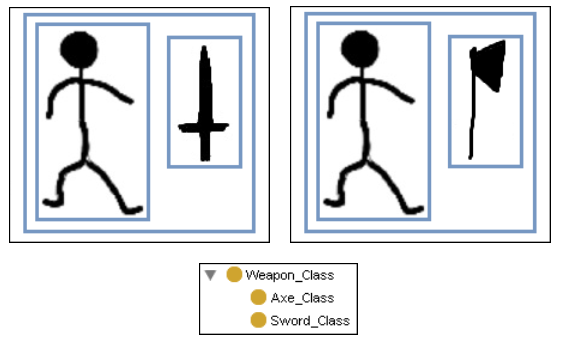}
& 

\begin{verbatim}
rule('Warrior', Scenes) :-
  scene(ID1, BB1, Class1, 'Human', Conf1, SS1),
  subclass_of('Weapon_Class', Class),
  interpretation(Class, Weapon),
  scene(ID2, BB2, Class2, Weapon, Conf2, SS2),
  append([scene(ID1, BB1, Class1, 'Human', Conf1, SS1)],
         [scene(ID2, BB2, Class2, Weapon, Conf2, SS2)],
         Scenes),
  relations(GR),
  jpl_call(GR, horizontal, [BB1, BB2], @(true)),
  jpl_call(GR, near, [BB1, BB2, 2.0], @(true)).
\end{verbatim}

\\
\bottomrule
\end{tabular}
\caption{Warrior Scene (Human + Weapon). Image and rule.}
\label{tab:WarriorScene2}
\end{scriptsize}
\end{table}

A composite scene can be defined by other composite scenes. For example, if we want to define a \texttt{Battle} scene as a combination of two composite \texttt{Warrior} scenes, a \texttt{rule} could be defined to check that two  \texttt{Warrior} scenes have been detected in the image, and that they are close enough. 
In general, the user can implement \texttt{rule} in any way, using all the expressive power of Prolog and creating auxiliary predicates for designing and implementing more complex rules. The rules presented so far only aggregate two scenes at a time, but of course it is possible to select a larger number.
For example, a scene of \texttt{War} could be formed by an arbitrary number of \texttt{Battle} scenes close to each other, as shown in the next paragraph.

\paragraph{Example 2: War scene (group of Battle).}

%\begin{figure}[h]
%\begin{center}
%\includegraphics[width=8cm]{images/65.png}
%\caption{War scene (group of Battle).}
%\label{fig:65}
%\end{center}
%\end{figure}  
%
Figure in Table \ref{tab:Battle2} shows a \texttt{War} scene consisting of three \texttt{Battle} scenes, close to each other. The \texttt{rule} implementation could be the one on the right of the table, which looks for all the asserted \texttt{Battle} scenes and nondeterministically selects some of them using the \texttt{sublist/2} predicate. Finally, it checks that those scenes are close enough to form a group (\texttt{jpl\_call(GR, group, [BBs, 10.0], @(true))}).
%
%\begin{verbatim}
%rule('War', Scenes) :-
%  findall(scene(ID, BB, Class, 'Battle', Conf, SS),
%          scene(ID, BB, Class, 'Battle', Conf, SS),
%          Battles),
%  sublist(Battles, Scenes),
%  relations(GR),
%  jpl_call(GR, group, [BBs, 10.0], @(true)).
%\end{verbatim}

\begin{table}[h]
\begin{scriptsize}
\begin{tabular}{m{5cm}m{8cm}}
\toprule
Example Image & Scene Interpretation Rule\\
\midrule
\includegraphics[width=5cm]{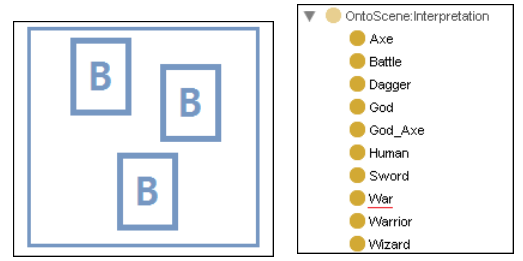}
& 

\begin{verbatim}
rule('War', Scenes) :-
  findall(scene(ID, BB, Class, 'Battle', Conf, SS),
          scene(ID, BB, Class, 'Battle', Conf, SS),
          Battles),
  sublist(Battles, Scenes),
  relations(GR),
  jpl_call(GR, group, [BBs, 10.0], @(true)).
\end{verbatim}

\\
\bottomrule
\end{tabular}
\caption{War scene (group of Battles).}
\label{tab:Battle2}
\end{scriptsize}
\end{table}

\subsection{Computing all the possible interpretations}
\label{knuthalgo}

The main functionality of \texttt{SceneInterpreter} consists of analyzing all the nodes in the scene graph to determine which of them can coexist in an interpretation (which has to contain all the basic scenes). Two nodes can coexist in the same interpretation if and only if they do not share any basic scene. For example, in the figure in Table \ref{tab:graph1}, node CS1 and node CS2 cannot coexist in an interpretation because they share BS2. 

This ``coexistence check'' resorts to the NP-complete exact cover problem \cite{DBLP:conf/coco/Karp72}. Let  X be the set of the basic scenes computed, and asserted, in the way discussed in Section \ref{serializing}. Each node in the scene graph identifies a subset of X: the scene graph is a collection S of subsets of a set X. By definition, an exact cover of X is a subcollection S* of S that satisfies two conditions:
\begin{enumerate}
\item    The intersection of any two distinct subsets in S* is empty, i.e., the subsets in S* are pairwise disjoint. In other words, each element in X is contained in at most one subset in S*.
\item    The union of the subsets in S* is X, i.e., the subsets in S* cover X. In other words, each element in X is contained in at least one subset in S*.
\end{enumerate}
A subcollection S* satisfying the two properties above is indeed what we name a {\em scene interpretation}.
\texttt{SceneInterpreter} implements Donald Knuth's {\em Algorithm X} for the exact cover problem \cite{DancingLinks00}. Algorithm X is a recursive, nondeterministic, depth-first, backtracking algorithm: the ideal algorithm for  Prolog!

If we disregard the code for managing matrices (an \texttt{update\_matrix} predicate is needed, whose code is not shown), the  Algorithm X' Prolog implementation is 14 lines long, excluding comments. 

The exact cover problem is represented in Algorithm X using a matrix A consisting of 0s and 1s. The goal is to select a subset of the rows so that the digit 1 appears in each column exactly once. 
Table \ref{tab:algoX} shows the Prolog code for the algorithm, implemented by the \texttt{knuth\_algo\_x} predicate:
\begin{verbatim}
knuth_algo_x/5.
knuth_algo_x(M, Nodes, NumNodes, AccSolution, Solution).
\end{verbatim}
\begin{itemize}
\item \texttt{M} represents the matrix associated with the collection S of subsets of X (which, in turn, is associated with the scene graph stored via the \texttt{image\_graph(G)} fact); it is represented in a standard way as a list of lists, making it possible to exploit the \texttt{transpose/2} predicate offered by the SWI-Prolog CLP(FD) library for  Constraint Logic Programming over Finite Domains\footnote{\url{https://www.swi-prolog.org/pldoc/doc/_SWI_/library/clp/clpfd.pl}, accessed on July 2019.}.  
\item \texttt{Nodes} is the list of nodes in the scene graph.
\item \texttt{NumNodes} is the number of nodes in the scene graph.
\item \texttt{AccSolution} is the accumulator argument.
\item \texttt{Solution} is unified with the solution, when the algorithm terminates.
\end{itemize}

\begin{table}[h]
{\small{
\begin{verbatim}
% if the matrix is empty, terminate by unifying the last argument
% with the accumulator
knuth_algo_x([], _, _, Solution, Solution) :- !.

% otherwise
knuth_algo_x(M, Nodes, NumNodes, AccSolution, Solution) :-

  % deterministically select one column containing as few 1s
  % as possible
  transpose(M, TM),
  get_col_with_less_ones(TM, NumNodes, 0, 0, ColIndex, ColCount),
  ColCount > 0,
  
  % non deterministically select one row with the selected 
  % column equal to 1
  member(Row, M),
  nth0(ColIndex, Row, 1, _),
  
  % update the partial solution
  nth0(Index, M, Row, _),
  nth0(Index, Nodes, SolutionNode, _),
  append(AccSolution, [SolutionNode], NewAccSolution),
  
  % update the matrix
  findall(I, nth0(I, Row, 1, _), ColumnsToRemove),
  update_matrix(M, ColumnsToRemove, NewM, RemovedRowsIndexes),
  
  % remove the nodes that were associated with the removed rows
  findall(N,(member(N,Nodes), nth0(I1, Nodes, N, _), 
  \+ member(I1, RemovedRowsIndexes)), NewNodes),
  
  % recursively call the algorithm on the reduced matrix and nodes
  knuth_algo_x(NewM, NewNodes, NumNodes, NewAccSolution, Solution).
\end{verbatim}
}}
\caption{Donald Knuth's {\em Algorithm X} implementation in Prolog.}
%\end{scriptsize}
\label{tab:algoX}
\end{table}

The nondeterministic choice of the row via the \texttt{member(Row, M)} goal allows the algorithm to ``clone'' itself into independent subalgorithms which work on a reduced version of the matrix \texttt{M}. Searching the state space is of course left to the Prolog interpreter. 

Each set of nodes in the graph scene which is an exact cover of the basic scenes is an interpretation of the input scene. The possible interpretations of the figure in Table \ref{tab:graph1}, left, are reported in the table right side.

%\begin{verbatim}
%I1 = [BS1, BS2, BS3, BS4 and BS5]
%I2 = [CS1, BS3, BS4, BS5]
%I3 = [CS1, BS3, CS3]
%I4 = [BS1, CS2, BS5]
%I5 = [BS1, CS4]
%\end{verbatim}

The set \texttt{[BS1, BS2, BS3]} is not an interpretation because it does not contain all the input images (BS4 and BS5 are missing) and \texttt{[CS1, CS4]} is not correct as well because both CS1 and CS4 share the same scene BS2 (and hence cannot coexist). 

\subsection{Filtering, sorting, and returning interpretations}
\label{filtering-sorting-returning}

Usually one input image generates many scene interpretations, some of which can be derived from others by substituting one aggregated scene with the sub-scenes which form it. \texttt{Sce\-ne\-In\-ter\-preter} can filter out interpretations that can be derived by others in this way. 
In the example above, I1 and I2 can be derived from I3 and can be filtered out: if we substitute CS1 with  its children BS1 and BS2, and (resp. or) CS3 with BS4 and BS5, we obtain I1 (resp. I2).

Each interpretation is checked against the others computed so far, to avoid duplicates due to the order of nodes in the interpretation, and is associated with a weight computed as the sum of the squares of the aggregated scenes lengths.
As an example, the weight of the following interpretations is 8 and 4, respectively.

\begin{verbatim}
I1 = [W1(Human-0, Sword-1), W2(Human-3, Sword-2)]. % Weight = 8
I2 = [Human-0, Sword-1, Human-3, Sword-2].         % Weight = 4
\end{verbatim}

Interpretations are sorted in decreasing weight order, from the one which aggregates more scenes together to the one where less aggregation rules have been exploited. In the example above, I1 ``aggregates more'' than I2 and comes before I2 in the list of computed interpretations, but both are returned. 

\subsection{SceneInterpreter at work}

In this section we show further examples in the Battle domain, each coming with an informal description and the interpretations that the Prolog interpreter generates when the \texttt{ge\-ne\-ra\-te\-All\-In\-ter\-pre\-ta\-tions} and \texttt{ge\-ne\-ra\-te\-Filtered\-In\-ter\-pre\-ta\-tions} predicates are called.
We consider images whose possible classifications are  \texttt{Human\_Class}, \texttt{Sword\_Class} and \texttt{Axe\_Class}, with maximum confidence. For each classification, we assume that only one interpretation exists:
\begin{verbatim}
interpretation('Human_Class', 'Human').
interpretation('Sword_Class', 'Sword').
interpretation('Axe_Class', 'Axe').
\end{verbatim}

%
%\begin{figure}[h]
%\begin{center}
%\includegraphics[width=6cm]{images/66bis.png}
%\caption{Input image.}
%\label{fig:66}
%\end{center}
%\end{figure}  

%Basic scenes generated by the images in Table \ref{fig:66} look like:
%\begin{verbatim}
%scene(ID1, BB1, 'Human_Class', 'Human', 1.0, [0]).
%scene(ID2, BB2, 'Sword_Class', 'Sword', 1.0, [1]).
%scene(ID3, BB3, 'Axe_Class', 'Axe', 1.0, [2]).
%\end{verbatim}

In Table \ref{tab:basicExample} the basic scenes generated for the images that will be used in the examples are reported.

\begin{table}[h]
\begin{scriptsize}
\begin{tabular}{m{4cm}m{9cm}}
\toprule
Input Image & Basic Scenes\\
\midrule
\includegraphics[width=4cm]{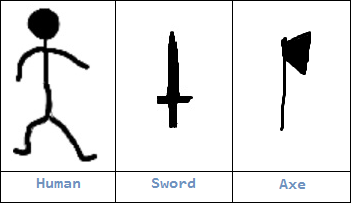} & 

\begin{verbatim}
scene(ID1, BB1, 'Human_Class', 'Human', 1.0, [0]).
scene(ID2, BB2, 'Sword_Class', 'Sword', 1.0, [1]).
scene(ID3, BB3, 'Axe_Class', 'Axe', 1.0, [2]).
\end{verbatim}
\\
\bottomrule
\end{tabular}
\caption{Basic scenes used in the next examples.}
\label{tab:basicExample}
\end{scriptsize}
\end{table}

In the next examples, for each scene we show the scene graph (generated by calling the \texttt{applyRules} method) and the generated interpretations. We consider the following composite scenes:\\
\texttt{Warrior = Human + Weapon} (\texttt{Sword} or \texttt{Axe})\\
with distance between the \texttt{BB} of \texttt{Human} and the \texttt{BB} of \texttt{Weapon} $\leq$ 2px.
\\
\texttt{   Battle = Warrior + Warrior}\\
with distance between the \texttt{BB}s $\leq$ 5px.\\

\paragraph{Example scene 1.}

%\begin{figure}[h]
%\begin{center}
%\includegraphics[width=3.2cm]{images/67.png}
%\caption{Example scene 1.}
%\label{fig:67}
%\end{center}
%\end{figure}

Table \ref{tab:SIExp1} shows a \texttt{Human} \texttt{(ID = 0)} close to another figure \texttt{(ID = 1)} that can be classified as \texttt{Sword\_Class} and \texttt{Axe\_Class}, and hence interpreted as \texttt{Sword} and \texttt{Axe}. 
The scene graph generated by  \texttt{applyRules} contains two  \texttt{Warrior}s, \texttt{W1} and \texttt{W2}. The generated interpretations are reported on the right of the table.

%Interpretations:
%\begin{verbatim}
%generateAllInterpretations (4):
%I1 = [W1(Human-0, Sword-1)].
%I2 = [W2(Human-0, Axe-1)].
%I3 = [Human-0, Sword-1].
%I4 = [Human-0, Axe-1].
%
%GenerateFilteredInterpretations (2):
%I1 = [W1(Human-0, Sword-1)].
%I2 = [W2(Human-0, Axe-1)].
%\end{verbatim}

\begin{table}[h]
\begin{scriptsize}
\begin{tabular}{m{4cm}m{8cm}}
\toprule
Input Image & Interpretations\\
\midrule
\includegraphics[width=2cm]{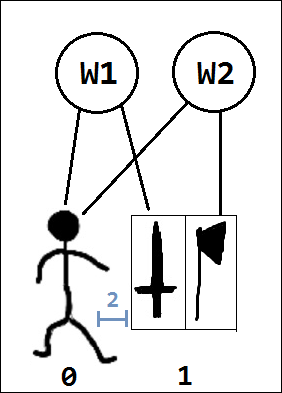} & 

\begin{verbatim}
generateAllInterpretations (4):
I1 = [W1(Human-0, Sword-1)].
I2 = [W2(Human-0, Axe-1)].
I3 = [Human-0, Sword-1].
I4 = [Human-0, Axe-1].

GenerateFilteredInterpretations (2):
I1 = [W1(Human-0, Sword-1)].
I2 = [W2(Human-0, Axe-1)].
\end{verbatim}
\\
\bottomrule
\end{tabular}
\caption{Example 1 of a complex scene with its interpretations.}
\label{tab:SIExp1}
\end{scriptsize}
\end{table}

\paragraph{Example scene 2.}

%\begin{figure}[h]
%\begin{center}
%\includegraphics[width=9cm]{images/68.png}
%\caption{Example scene 2.}
%\label{fig:68}
%\end{center}
%\end{figure}

Table \ref{tab:SIExmp2} shows four \texttt{Human}s and four \texttt{Weapon}s. Each \texttt{Human} is close enough to the \texttt{Weapon} at its right to be interpreted as a  \texttt{Warrior} (\texttt{W1, W2, W3, W4}), and each \texttt{Warrior} is close enough to the adjacent \texttt{Warrior} to be considered as a  \texttt{Battle} (\texttt{B1, B2, B3}). The first five generated interpretations, on a total of 29 ones, are reported on the right of the table.

%Interpretations:
%\begin{verbatim}
%generateAllInterpretations (29):
%I1 = [B1(W1, W2), B3(W3, W4)].
%I2 = [W1(Human-0, Sword-1), B2(W2, W3), W4(Human-6, Axe-7)].
%I3 = [W1(Human-0, Sword-1), W2(Human-2, Sword-3), B3(W3, W4)].
%I4 = [B1(W1, W2), W3(Human-4, Sword-5), W4(Human-6, Axe-7))].
%I5 = [Human-0, Sword-1, B2(W2, W3), W4(Human-6, Axe-7)].
%\end{verbatim}
%(24 more interpretations are generated).
%
%\begin{verbatim}
%GenerateFilteredInterpretations (2):
%I1 = [B1(W1, W2), B3(W3, W4)].
%I2 = [W1(Human-0, Sword-1), B2(W2, W3), W4(Human-6, Axe-7)].
%\end{verbatim}
%

\begin{table}[h]
\begin{scriptsize}
\begin{tabular}{m{5.5cm}m{7.5cm}}
\toprule
Input Image & Interpretations\\
\midrule
\includegraphics[width=5.4cm]{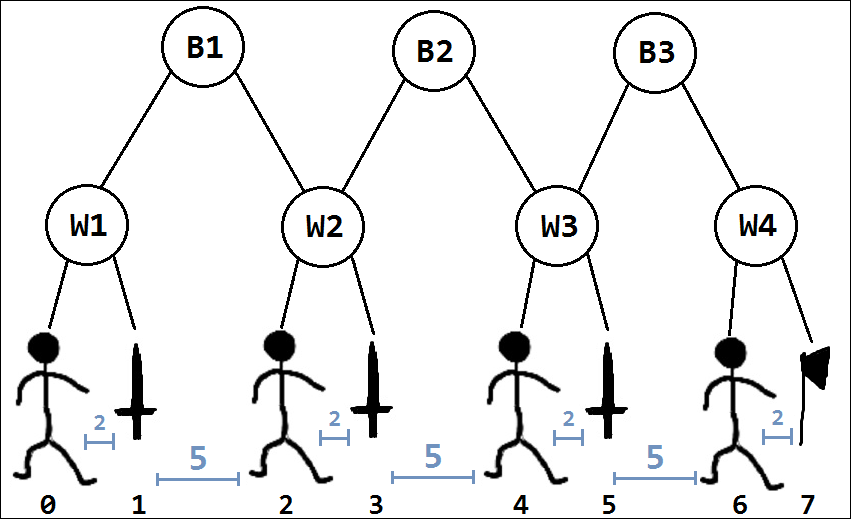} & 

\begin{verbatim}
generateAllInterpretations (29):
I1 = [B1(W1, W2), B3(W3, W4)].
I2 = [W1(Human-0, Sword-1), B2(W2, W3), 
      W4(Human-6, Axe-7)].
I3 = [W1(Human-0, Sword-1), W2(Human-2, Sword-3), 
      B3(W3, W4)].
I4 = [B1(W1, W2), W3(Human-4, Sword-5), 
      W4(Human-6, Axe-7))].
I5 = [Human-0, Sword-1, B2(W2, W3), 
      W4(Human-6, Axe-7)].
\end{verbatim}
(24 more interpretations are generated).

\begin{verbatim}
GenerateFilteredInterpretations (2):
I1 = [B1(W1, W2), B3(W3, W4)].
I2 = [W1(Human-0, Sword-1), B2(W2, W3), 
      W4(Human-6, Axe-7)].
\end{verbatim}
\\
\bottomrule
\end{tabular}
\caption{Example 2 of a complex scene with its interpretations.}
\label{tab:SIExmp2}
\end{scriptsize}
\end{table}

\vspace{-0.5cm}

\paragraph{Example scene 3.}

%\begin{figure}[h]
%\begin{center}
%\includegraphics[width=5cm]{images/69.png}
%\caption{Example scene 3.}
%\label{fig:69}
%\end{center}
%\end{figure}

Table \ref{tab:SIExmp3} shows two \texttt{Human}s on the right and on the left of the picture, both close to the two \texttt{Sword}s in the center. Each \texttt{Human} can only be associated with the \texttt{Sword} that is closest to him (\texttt{Human} 0 cannot be associated with \texttt{Sword} 2 and the same for 3 and 1). Hence, the only possible interpretations are two \texttt{Warrior}s \texttt{W1} and \texttt{W2} and one \texttt{Battle B1}. The generated interpretations are reported on the right of the table.

%Interpretations:
%\begin{verbatim}
%generateAllInterpretations (3):
%I1 = [B1(W1, W2)].
%I2 = [W1(Human-0, Sword-1), W2(Human-3, Sword-2)].
%I3 = [Human-0, Sword-1, Human-3, Sword-2].
%
%GenerateFilteredInterpretations (1):
%I1 = [B1(W1, W2)].
%\end{verbatim}

\begin{table}[h]
\begin{scriptsize}
\begin{tabular}{m{5cm}m{8cm}}
\toprule
Input Image & Interpretations\\
\midrule
\includegraphics[width=3cm]{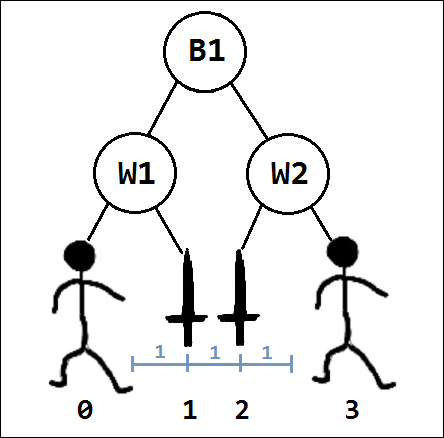} & 

\begin{verbatim}
generateAllInterpretations (3):
I1 = [B1(W1, W2)].
I2 = [W1(Human-0, Sword-1), W2(Human-3, Sword-2)].
I3 = [Human-0, Sword-1, Human-3, Sword-2].

GenerateFilteredInterpretations (1):
I1 = [B1(W1, W2)].
\end{verbatim}
\\
\bottomrule
\end{tabular}
\caption{Example 3 of a complex scene with its interpretations.}
\label{tab:SIExmp3}
\end{scriptsize}

\end{table}

\newpage

\section{Case Study: Interpreting Scenes from the Rock Art Domain}
\label{sec:CaseStudy}

In this section, we present OntoScene at work. The domain where we experimented  it is the one introduced in Section \ref{initscen}: Mount Bego's prehistoric rock art.

\subsection{Studies by Clarence Bicknell and Henry de Lumley}

Archaeologists and historians look at the area around Mount Bego as an incredibly valuable source of knowledge, due to the up to 40,000 figurative petroglyphs and 60,000 non-figurative petroglyphs scattered over a large area at an altitude of 2,000 to 2,700 meters.

The historical relevance of the Mount Bego petroglyphs is unquestionable, as they date back to the early Bronze Age, when humans left no written evidences and the only witnesses of their existence are their tools and, indeed, their drawings.

The explorer who first realized the importance of Mount Bego carvings was Clarence Bicknell,  who, at the turn of the 20th century, created an important catalogue of most of the petroglyphs in Mount Bego \cite{Bicknell13}.  

Many years after Bicknell's campaigns, several teams led by Henry de Lumley have been surveying and mapping this archaeological area starting from 1967 \cite{Bianchi,Lumley}.

The University of Genova owns a collection of 16,000 drawings and reliefs made by Clarence Bicknell between 1898 and 1910, in his campaigns on Mount Bego. Bicknell's Legacy also includes nine notebooks, filled with notes in Victorian English, mostly unpublished. The publication on the web of about 350 images from the Bicknell's drawings and reliefs (Rolls 8, 20, 23, available on the Bicknell Legacy web site) along with their classification was one of the results of the IndianaMAS research project.

The images used for the experiments presented in this section and in the Appendix come from the Bicknell's Legacy and from \begin{TBF} the book by de Lumley and Echassoux \cite{Bego}\end{TBF}: we report an identifier under each image to refer to the first (abbreviated into BL, R. for Roll and P. for page) or to the second (abbreviated into DE, P. for page and F. for figure number).

For each type of scene in the dataset, three or four images were manually selected to represent the most frequent recognized patterns. 
The \texttt{Detector} and \texttt{Classifier} modules were simulated by manually drawing BBs around the sub-images of the scene and assigning them the classifications provided by Dr. Nicoletta Bianchi, who collaborated with us in the IndianaMAS project and in the construction of the Bicknell Legacy website. With her help, we also produced a natural language interpretation rule for Bicknell's images and we translated them in Prolog for each scene type. As far as de Lumley and Echassoux' images are concerned, the natural language interpretation rules are those written in their book.

\subsection{Experiments}

% To perform the experiments, a Prolog program has been developed to load an input scene modeled in Prolog as a list of \verb|image/3| facts (representing the input images): each \texttt{image(ID, BB, Classifications)} fact represents one of the sub-images or tokens in the image that should be interpreted, where  \texttt{ID} represents the identifier of an Image, \texttt{BB} is a term of type \verb|bb(X,Y,W,H)| with \texttt{X} and \texttt{Y} representing the coordinates of the top left point of the BB associated with the image and \texttt{W} and \texttt{H} representing its width (Width) and height (Height), and \texttt{Classifications} is a list of terms of the type \texttt{class(Class, Conf)} with \texttt{Class} representing a classification consistent with the domain ontology and \texttt{Conf} the associated confidence. 

We analyzed 34 images of scenes, covering 9 different interpretations.
In the sequel we report the facts and rules used to interpret the pastoral scene, and the results of the performed tests; to make the paper more compact, for three more scenes we only provide a textual explanation of the scene interpretation and the computed results. The Prolog rules for these three scenes can be found in the Appendix, along with five more examples. 
For sake of clarity, the \verb|bb(X,Y,W,H)| argument of the \texttt{image} predicate is omitted in the following tables, which report the selected images and the respective interpretations with the test results.

\subsection{Pastoral scene (corniforms group)}
\label{corniformgroup}

\noindent {\bf Interpretation of the scene by archaeologists}: A group of corniforms close to each others represents a pastoral scene.\\

%The BBs should be near or partially overlapping.

\noindent {\bf Association between sub-image classification and sub-image interpretation}: 
{\small{
\begin{verbatim}
interpretation('Corniform_Class', 'Corniform').
\end{verbatim}
}}

\noindent {\bf Rules for scene interpretation}:
{\small{
\begin{verbatim}
rule('Group_Of_Corniforms', Scenes) :-
   findall(scene(ID, BB, Cl, 'Corniform', Conf, SS),
           scene(ID, BB, Cl, 'Corniform', Conf, SS),
           Corns),
   sublist(Corns, Scenes),
   findall(BB, 
           member(scene(_, BB, _, _, _, _), Scenes), 
           BBs),
   prolog_list_to_java_list(BBs, JavaBBs),
   relations(GR),
   jpl_call(GR, group, [JavaBBs, 0.5], @(true)).
\end{verbatim}
}}

\noindent {\bf Explanation}: the rule 
\begin{itemize}
\item creates the set of corniforms in the scene by calling \texttt{findall(scene(ID, BB, Cl, 'Corniform', Conf, SS),
           scene(ID, BB, Cl, 'Corniform', Conf, SS),
           Corns))},
\item non deterministically picks one partition of the set of corniforms by calling \texttt{sublist(Corns, Scenes)}, 
\item for the selected partition, retrieves the list of bounding boxes of the images therein by calling \texttt{findall(BB, 
           member(scene(\_, BB, \_, \_, \_, \_), Scenes), 
           BBs)}, 
\item transforms the Prolog list \texttt{BBs} into a format suitable for being passed as an argument to a Java call (\texttt{prolog\_list\_to\_java\_list(BBs, JavaBBs)}, and finally 
\item checks if the bounding boxes form a group by calling 
               \texttt{relations(GR), jpl\_call(GR, group, [JavaBBs, 0.5], @(true))}.
\end{itemize}

\noindent Table \ref{tab:pastoral} reports the results of the four analyzed images, all correctly interpreted.

\begin{table}[h]
\begin{tiny}
\begin{tabular}{m{1.6cm}m{5cm}m{5.5cm}m{0.8cm}}
\toprule
Image & Input Single Images & Resulting Interpretation & Final result\\
\midrule
\includegraphics[width=1.3cm]{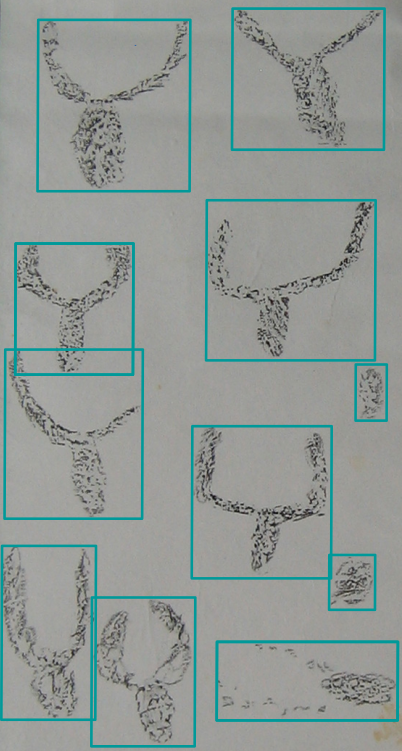} BL, R. 20, P. 5 & 

\verb|image(0, [class('Corniform_Class', 1.0)])|

\verb|image(1, [class('Corniform_Class', 1.0)])|

\verb|image(2, [class('Corniform_Class', 1.0)])|

\verb|image(3, [class('Corniform_Class', 1.0)])|

\verb|image(4, [class('Corniform_Class', 1.0)])|

\verb|image(5, [class('Corniform_Class', 1.0)])|

\verb|image(6, [class('Corniform_Class', 1.0)])|

\verb|image(7, [class('Solid_Ellipse_Class', 1.0)])|

\verb|image(8, [class('Corniform_Class', 1.0)])|

\verb|image(9, [class('Corniform_Class', 1.0)])|

\verb|image(10, [class('Solid_Ellipse_Class', 1.0)])|

\verb|image(11, [class('Corniform_Class', 0.6)])|
& 

\verb|I1 = [Group_Of_Corniforms(Corniform-0, Corniform-1,|

\verb|Corniform-2, Corniform-3, Corniform-4, Corniform-5,|

\verb|Corniform-6, Corniform-8, Corniform-9, Corniform-11)|,

\verb|Cup_Stone_7, Cup_Stone_10]|
                          
\verb|I2 = [Corniform-0, Corniform-1, Corniform-2, ...,|

\verb|Cup_Stone_7, Cup_Stone_10]| & Passed\\

\includegraphics[width=1.3cm]{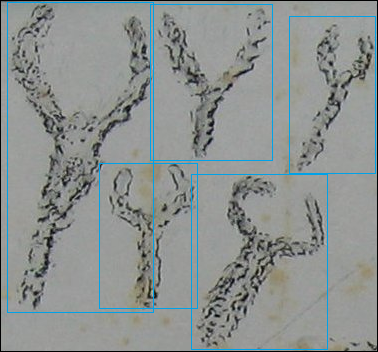} BL, R. 20, P. 63  & 

\verb|image(0, [class('Corniform_Class', 1.0)])|

\verb|image(1, [class('Corniform_Class', 1.0)])|

\verb|image(2, [class('Corniform_Class', 1.0)])|

\verb|image(3, [class('Corniform_Class', 1.0)])|

\verb|image(4, [class('Corniform_Class', 1.0)])| & 

\verb|I1 = [Group_Of_Corniforms(Corniform-0, Corniform-1,|

\verb|Corniform-2, Corniform-3, Corniform-4)]|
                          
\verb|I2 = [Corniform-0, Corniform-1, Corniform-2,|

\verb|Corniform-3, Corniform-4]| & Passed\\

\includegraphics[width=1.3cm]{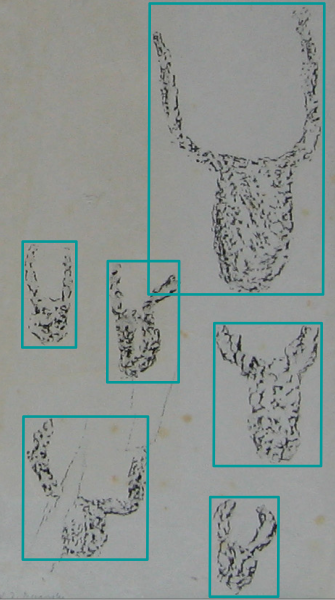} BL, R. 20, P. 80  &

\verb|image(0, [class('Corniform_Class', 1.0)])|

\verb|image(1, [class('Corniform_Class', 1.0)])|

\verb|image(2, [class('Corniform_Class', 1.0)])|

\verb|image(3, [class('Corniform_Class', 1.0)])|

\verb|image(4, [class('Corniform_Class', 0.7)])|

\verb|image(5, [class('Corniform_Class', 1.0)])])|
& 

\verb|I1 = [Group_Of_Corniforms(Corniform-0, Corniform-1,|

\verb|Corniform-2, Corniform-3, Corniform-4, Corniform-5)]|
                          
\verb|I2 = [Corniform-0, Corniform-1, Corniform-2, ...]| & Passed\\

\includegraphics[width=1.3cm]{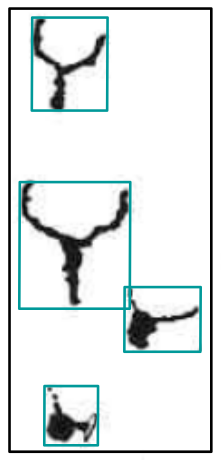} DL, P. 230, F. 202  &

\verb|image(0, [class('Corniform_Class', 1.0)])|

\verb|image(1, [class('Corniform_Class', 1.0)])|

\verb|image(2, [class('Corniform_Class', 1.0)])|

\verb|image(3, [class('Corniform_Class', 1.0)])|
& 

\verb|I1 = [Group_Of_Corniforms(Corniform-0, Corniform-1,|

\verb|Corniform-2, Corniform-3)]|
                          
\verb|I2 = [Corniform-0, Corniform-1, Corniform-2,| 

\verb|Corniform-3]|& Passed\\

% \includegraphics[width=1.3cm]{images/bl2302.png} BL, R. 23, P. 2  & 
% 
% \verb|image(0, [class('Corniform_Class', 1.0)])|
% 
% \verb|image(1, [class('Corniform_Class', 1.0)])|
% 
% \verb|image(2, [class('Corniform_Class', 1.0)])|
% 
% \verb|image(3, [class('Corniform_Class', 1.0)])|
% 
% \verb|image(4, [class('Corniform_Class', 1.0)])|
% 
% \verb|image(5, [class('Empty_Ellipse_Class', 1.0)])|
% 
% \verb|image(6, [class('Corniform_Class', 1.0)])|
% 
% \verb|image(7, [class('Corniform_Class', 0.6)])|
% & 
% 
% \verb|I1 = [Group_Of_Corniforms(Corniform-0, Corniform-1,|
% 
% \verb|Corniform-2, Corniform-3, Corniform-4, Corniform-6,|
% 
% \verb|Corniform-7), Geometric_Figure_5]|
                          % 
% \verb|I2 = [Corniform-0, Corniform-1, Corniform-2, ...,|
% 
% \verb|Geometric_Figure_5]| & Passed\\

\bottomrule

\end{tabular}
\caption{Results of the interpretation of Pastoral scenes.}
\label{tab:pastoral}
\end{tiny}
\end{table}

\subsection{Ritual sacrifice}

\noindent {\bf Interpretation of the scene by archaeologists}: One halberd near one, or few more, corniforms, represents a ritual sacrifice.
From the analysis of the available images, we identified three patterns: one where the BB of the corniform is inside the one of the halberd, another one where the two BBs are overlapping, and a last one where there are more corniforms.\\

\noindent {\bf Explanation}: the rule shown in Section \ref{appendix:rs} selects one halberd and another scene called \verb|Victim| (a corniform or a group of corniforms) in the \verb|Scenes| list. The check succeeds if the halberd's BB contains or overlaps with the one of the \verb|Victim|.\\

\noindent Table \ref{tab:ritSac} reports the results of the four analyzed images: the last one has not been recognized because the two bounding boxes are neither overlapping nor one inside the other, as required by the rule.

\begin{table}[h]
\begin{tiny}
\begin{tabular}{m{2.1cm}m{4.3cm}m{5.5cm}m{0.8cm}}
\toprule
Image & Input Single Images & Resulting Interpretation & Final result\\
\midrule
\parbox{2.1cm}{\includegraphics[width=1.3cm]{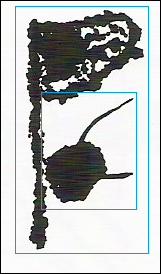}\\ DE, P. 187, F. 154(3)} & 

\verb|image(0,[('Halberd_Class',1.0)])|

\verb|image(1,[('Corniform_Class',1.0)])| & 

\verb|I1=[Ritual_Sacrifice(Halberd-0,Corn-1)]|

\verb|I2=[Halberd-0, Corn-1]|& Passed\\

\parbox{2.1cm}{\includegraphics[width=1.3cm]{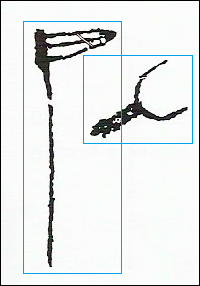}\\ DE, P. 187, F. 154(10)}  & 

\verb|image(0,[('Halberd_Class',1.0)])|

\verb|image(1,[('Corniform_Class',1.0)])| & 

\verb|I1=[Ritual_Sacrifice(Halberd-0,Corn-1)]|

\verb|I2=[Halberd-0, Corn-1]|& Passed\\

\parbox{2.1cm}{\includegraphics[width=1.3cm]{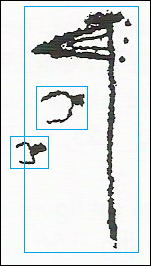}\\ DE, P. 187, F. 154(12)}  &

\verb|image(0,[('Halberd_Class',1.0)])|

\verb|image(1,[('Corniform_Class',1.0)])| 

\verb|image(2,[('Corniform_Class',1.0)])|& 

\verb|I1=[Ritual_Sacrifice(Halberd-0,|

\verb|Group_Of_Corniforms(Corniform-1,Corniform-2))]|

\verb|I2=[Halberd-0,|

\verb|Group_Of_Corniforms(Corn-1, Corn-2)]|

\verb|I3=[Halberd-0, Corn-1, Corn-2]| & Passed\\

% \includegraphics[width=1.3cm]{images/80.png} DE, P. 187, F. 154(7)  & 
% 
% \verb|image(0,[('Halberd_Class',1.0)])|
% 
% \verb|image(1,[('Corniform_Class',1.0)])| & 
% 
% \verb|I1=[Halberd-0,Corniform-1]| & Failed (Disjoint BBs)\\

\parbox{2.1cm}{\includegraphics[width=2cm]{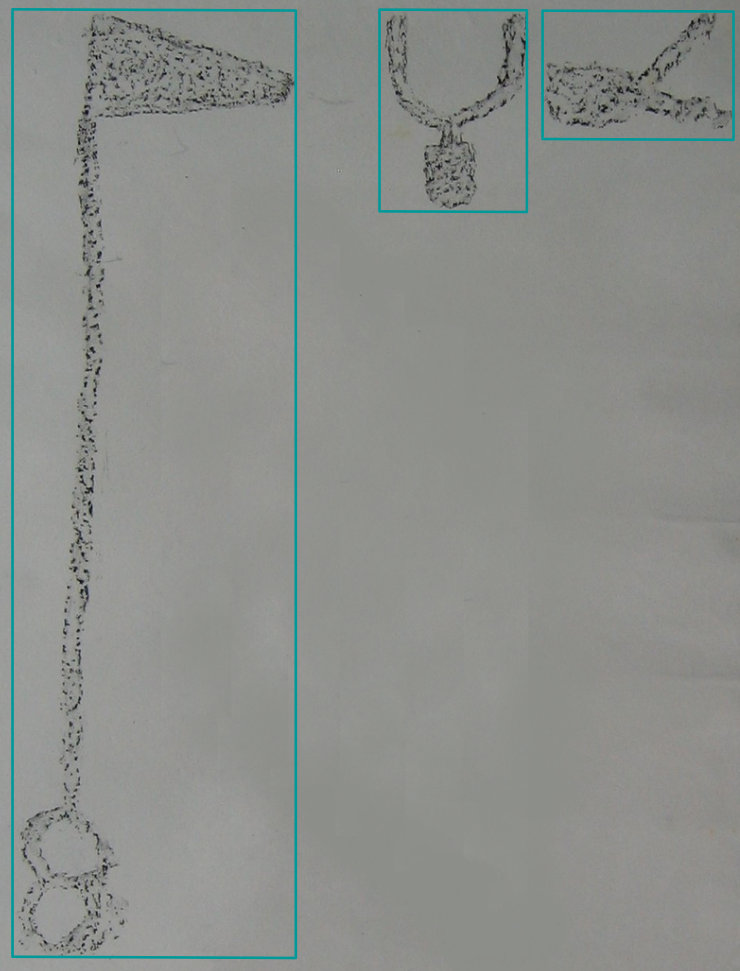}\\ BL, R. 8, P. 165}  &

\verb|image(0,[('Halberd_Class',1.0)])|

\verb|image(1,[('Corniform_Class',1.0)])| 

\verb|image(2,[('Corniform_Class',1.0)])|& 

\verb|I1=[Halberd-0,|

\verb|Group_Of_Corniforms(Corn-1, Corn-2)]|

\verb|I2=[Halberd-0, Corn-1, Corn-2]| & Failed (no overlap)\\

\bottomrule

\end{tabular}
\caption{Results of the interpretation of Ritual Sacrifice scenes.}
\label{tab:ritSac}
\end{tiny}
\end{table}

\subsection{Bull God birth}
%%%%%%%%%%%%%%%%%%%%%%%%%%%%%%%%%%%%%%%%%%%%%%%%%%%%%%%%%%%

\begin{figure}[h]
\begin{center}
\includegraphics[width=1.8cm]{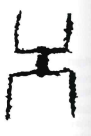} 
\caption{The High Goddess, DE, P. 328, F. 342.}
\label{fig:highgoddess}
\end{center}
\end{figure}

\noindent {\bf Interpretation of the scene by archaeologists}: One corniform below the High Goddess, shown in Figure \ref{fig:highgoddess}, represents the Bull God born by the High Goddess.
By analyzing the available images, two patterns were discovered: the first is where the high Goddess is above the Bull God, and close to him; the other is where she is above and partially overlaps with him.\\

\noindent {\bf Explanation}: the rule shown in Section \ref{appendix:bgb}  checks whether the token recognized as High Goddess is vertically aligned with the token representing the Bull God, and either overlaps with it, or it is close to it.\\

\noindent Table \ref{tab:bullgodbirth} reports the results of the four analyzed images:  the fourth one has not been correctly interpreted because the High Goddess is not close enough to the Bull God. The problem might be easily solved by changing the proximity parameter in  \texttt{jpl\_call(GR, near, [BB1, BB2, 0.5], @(true))}  from \texttt{0.5} to a higher value. Nevertheless, given that in most scenes representing the Bull God birth, the High Goddess is very close to him, increasing the proximity threshold might cause scenes with the High Goddess and one unrelated corniform nearby to be interpreted in the wrong way.

\begin{table}[h!]
\begin{tiny}
\begin{tabular}{m{1.5cm}m{5cm}m{5cm}m{1.2cm}}
\toprule
Image & Input Single Images & Resulting Interpretation & Final result\\
\midrule
\includegraphics[width=1.3cm]{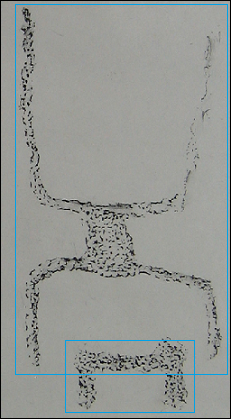} BL, R. 20, P. 134 & 

\verb|image(0, [class('Double_Appendixes', 1.0)])|

\verb|image(1, [class('Corniform_Class', 1.0)])| & 

\verb|I1 = [HG_Giving_Birth_BG(High_Goddess-0,|

\verb|Bull_God-1)]|

\verb|I2 = [High_Goddess-0, Bull_God-1]| & Passed\\

\includegraphics[width=1.3cm]{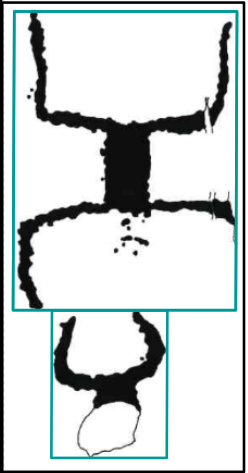} DE, P. 330, F. 345(2)  & 

\verb|image(0, [class('Double_Appendixes', 1.0)])|

\verb|image(1, [class('Corniform_Class', 1.0)])| & 

\verb|I1 = [HG_Giving_Birth_BG(High_Goddess-0,|

\verb|Bull_God-1)]|

\verb|I2 = [High_Goddess-0, Bull_God-1]| & Passed\\

\includegraphics[width=1.3cm]{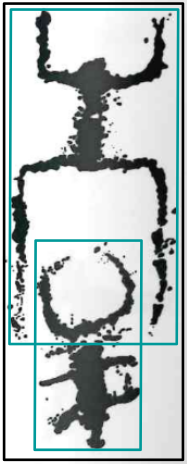} DE, P. 330, F. 345(4)  & 

\verb|image(0, [class('Double_Appendixes', 1.0)])|

\verb|image(1, [class('Corniform_Class', 1.0)])| & 

\verb|I1 = [HG_Giving_Birth_BG(High_Goddess-0,|

\verb|Bull_God-1)]|

\verb|I2 = [High_Goddess-0, Bull_God-1]| & Passed\\

\includegraphics[width=1.3cm]{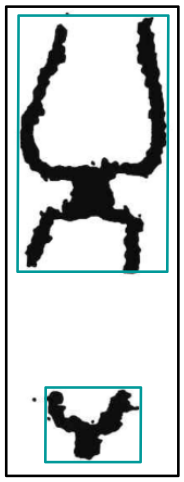} DE, P. 330, F. 345(3)  &

\verb|image(0, [class('Double_Appendixes', 1.0)])|

\verb|image(1, [class('Corniform_Class', 1.0)])| & 

\verb|I1 = [High_Goddess-0, Bull_God-1]| & Failed (not close enough)\\

\bottomrule

\end{tabular}
\caption{Results of the interpretation of Bull God Birth scenes.}
\label{tab:bullgodbirth}
\end{tiny}
\end{table}

\subsection{Storm God}

\noindent {\bf Interpretation of the scene by archaeologists}: One dagger and one reticulum with some overlaps represent the Storm God.\\

\noindent {\bf Explanation}:  the rule shown in Section \ref{appendix:sg}  searches for a dagger and a reticulum, checking if they overlap.\\

\noindent Table \ref{tab:storm} reports the results of four analyzed images; the first three ones have been correctly interpreted. The last one has not, because the reticulate and the dagger are very close, but do not overlap. 

\begin{table}[!htb]
\begin{tiny}
\begin{tabular}{m{2.3cm}m{4.4cm}m{5.3cm}m{0.8cm}}
\toprule
Image & Input Single Images & Resulting Interpretation & Final result\\
\midrule
\includegraphics[width=1.3cm]{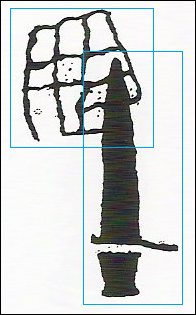} DE, P. 171, F. 133(2) & 

\verb|image(0,[('Dagger_Class',1.0)])|

\verb|image(1,[('Reticulum_Class',1.0)])| & 

\verb|I1=[Storm_God(Dagger-0, Reticulum-1)]|

\verb|I2=[Dagger-0, Reticulum-1]|& Passed\\

\includegraphics[width=1.3cm]{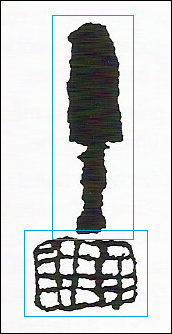} DE, P. 171, F. 133(4)  & 

\verb|image(0,[('Dagger_Class',1.0)])|

\verb|image(1,[('Reticulum_Class',1.0)])| & 

\verb|I1=[Storm_God(Dagger-0, Reticulum-1)]|

\verb|I2=[Dagger-0, Reticulum-1]|& Passed\\

\includegraphics[width=1.3cm]{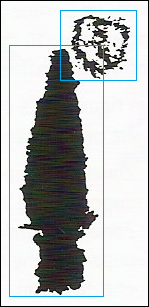} DE, P. 171, F. 133(5)  &

\verb|image(0,[('Dagger_Class',1.0)])|

\verb|image(1,[('Reticulum_Class',1.0)])| & 

\verb|I1=[Storm_God(Dagger-0, Reticulum-1)]|

\verb|I2=[Dagger-0, Reticulum-1]|& Passed\\

\includegraphics[width=2.2cm]{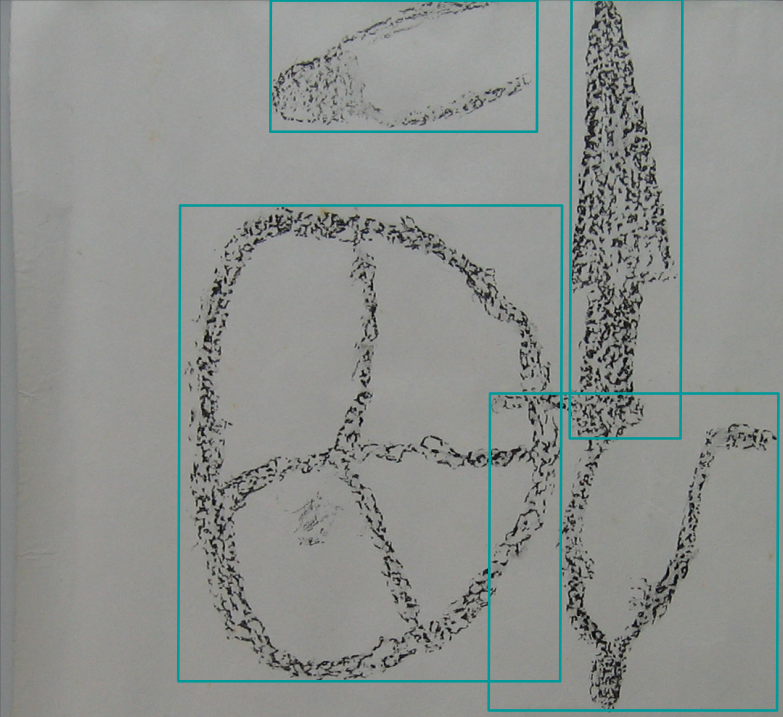} BL, R. 20, P. 33  &

\verb|image(0,[('Corniform_Class',1.0)])|

\verb|image(1,[('Reticulum_Class',1.0)])|

\verb|image(2,[('Dagger_Class',1.0)])|

\verb|image(3,[('Corniform_Class',1.0)])| & 

\verb|I1=[Dagger-2, Reticulum-1,Corn-0, Corn-3]|& Failed (no overlap)\\

\bottomrule

\end{tabular}
\caption{Results of the interpretation of Storm God scenes.}
\label{tab:storm}
\end{tiny}
\end{table}

\subsection{Discussion}

\paragraph{Suitability of Prolog for modelling and implementing the scene interpretation rules.} The power of Prolog for specifying scene interpretation rules is properly exemplified by the rule in Section \ref{corniformgroup} that exploits the \texttt{findall} all-solutions predicate for collecting all the images interpreted as corniforms into one set, generates one partition of the set in a nondeterministic way, and tests whether this partition enjoys the definition of being a group. If it does not, another partition is generated in backtracking and tested. By putting the \texttt{sublist} predicate inside a \texttt{findall} one, and then running the ``is a group?'' test on all the computed solutions, we would have obtained many more interpretations, one for each sub-group of corniforms in the scene. To keep it as simple and efficient as possible, the \texttt{rule('Group\_Of\_Corniforms', Scenes)} goal  succeeds as soon as the first group is found.  
While this rule was directly implemented in Prolog by the authors, based on the trivial intuition of what is a group of corniforms, other rules where sketched by the domain expert Dr. Nicoletta Bianchi using the formalism presented in Table \ref{tab:BNF}, and then translated by the authors into Prolog, following translation rules that can be easily automatized. This is the case, for example, of the Bull God birth rule presented in Section \ref{appendix:bgb}, whose rule in the user-friendly syntax is
{\small{
\begin{verbatim}
rule('HG_Giving_Birth_BG', [X,Y]) {	              
  High_Goddess(X);                                   
  Bull_God(Y);                                   
  (vertical(X,Y) or near(X,Y) or overlap(X,Y))                            
}                                               
\end{verbatim}
}}

\paragraph{Test results.}  We consider one test passed when OntoScene returns the correct interpretation, possibly together with other ones; 29 scenes out of 34 were correctly interpreted. 
The 5 scenes whose interpretation failed, did not satisfy the geometric constraints that the associated rule imposed. Failures are due to sub-images in the scene which do not overlap, while they should according to the rule, or that are not close enough, or that do not respect the expected orientation. In one case, failure is due to the lack of a suitable implementation for a geometric relation, ``around''. Given that scenes in this domain present a high variability, even when they have been resorted to the same interpretation by the domain experts,  writing ``the perfect rules'', keeping them as compact as possible, and as few as possible, is very hard. For example, the last test presented in Section \ref{appendix:arit} fails because the priest is above the repository, whereas the rule designed by the expert only accepts scenes where the priest is (or the two priests are) below. 
Adding one rule for coping with the failed test would not be difficult, but Nicoletta Bianchi knew that scenes like the one that failed the test are definitely less frequent than those that passed the test, and she suggested that -- in some cases -- obtaining a false negative could be better than designing many complex rules. In fact, OntoScene is meant to be a support to the domain experts, and not to substitute them in any way. Having a sound tool as OntoScene is, allows the expert to trust the ``Passed'' result, and to check only the ``Failed'' one. Although the human in the loop is still required, this approach may save a lot of time.

\paragraph{Likelihood of interpretations.} \verb|SceneInterpreter| computes all the scene interpretations which are consistent with the provided rules, but says nothing on the likelihood of one interpretation versus another. Coping with this further refinement does not represent a technical obstacle as it just resorts to sorting the elements in the list of computed interpretations according to some criterion. The actual obstacle is eliciting the sorting criterion from the domain experts, and formalizing it.
In all the 29 passed tests, the first interpretation returned, namely the one which ``aggregates more''  (see Section \ref{filtering-sorting-returning}), turned out to be the correct one. 
This observation might suggest some heuristics for pruning the search tree, such as keeping the weight of the best interpretation obtained so far, and avoiding to expand branches whose weight is expected to be lower. 
However, the fact that this simple sorting criterion worked finely in the rock art application domain, tells nothing on its generality. 
Different domain experts may have different personal opinions on how to select the correct interpretation of a scene, among many plausible ones, and associating a likelihood weight with each scene is not only domain dependent, but even domain expert dependent. This makes general and universally accepted sorting criteria difficult to assess: we did not face this issue in this paper, but it could be addressed either by integrating a heuristic criterion in the Algorithm X presented in Section \ref{knuthalgo} to stop recursion before the matrix \texttt{M} is empty, or by adding a post-processing stage of the \verb|SceneInterpreter| output into the framework data flow.
In the first case, the solution could be computed more efficiently, but could even get lost if the heuristic is not precise enough. In the second case, all the solutions should be computed, and efficiency would not benefit from the post-processing. 

%The experiments are meant to demonstrate the potential of the OntoScene framework, not to evaluate the precision of the used rules. %rather than to test its functionalities in an exhaustive way. To this aim, we also include rules that fail to interpret some scenes, and that could be easily modified in order to succeed in that specific interpretation, but whose modification could generate false positives on other scenes.  

\paragraph{Performance.} We did not assess the performance of OntoScene, both because efficiency was not our main concern, and because our experiments were run on scenes with no more than 11 sub-images: too few to raise efficiency issues. Despite the implemented optimization of Donald Knuth's Algorithm X, where selection of the column to remove is made in a clever way, the complexity of the problem itself is high, and the only way to reduce it would be to give up finding the exact solution, and integrate some heuristics in the algorithm.

If stress-tested on scenes consisting of a large number of sub-images, we expect that OntoScene bottleneck should turn out to be \verb|SceneInterpreter|, which would be a bottleneck even if implemented in any other language, because of the complexity of the exact cover algorithm it implements. \begin{TBF}Dovier et al. \cite{DBLP:conf/iclp/DovierFP05}\end{TBF} show how different NP-complete problems could be solved with either ASP \cite{lifschitz1999answer} or CLP(FD) \cite{marriott}, also on inputs with size greater than 2000. Based on these results, and considering that they date back to 15 years ago, we may suppose that, with today's computing power, with efficient Prolog implementations, and possibly with a careful exploitation of more advanced technologies like ASP and CLP(FD), we could use \verb|SceneInterpreter| on scenes with 2000 sub-images or more.  

We point out, however, that adopting OntoScene to model scenes with hundreds or thousands of sub-images does not seem a viable approach to scene interpretation, and not because of performance issues. Rule modelling is worth the effort if the modelled rules are general enough to cover a large number of different scenes, but the more the scene elements, the more specific the rule. For example, designing an OntoScene rule for interpreting the scene represented in the Parthenon frieze would require to model the relations holding between/among 378 human figures and 245 animals. A precise rule for achieving this goal would succeed on the Parthenon frieze, and would fail on anything else, and its usefulness would be very limited.

\section{Conclusions and Future Works}\label{sec:concl-future}

OntoScene is a modular platform aimed at supporting the interpretation of complex scenes based on ontologies and logical rules defined in Prolog. Ontologies allow the designer to formalize the domain and make the system modular and interoperable with existing MASs, while Prolog provides a solid basis to define complex rules of interpretation in a way that can be affordable even for people with no background in Computational Logics. The feedback we got from Nicoletta Bianchi, with whom we designed the rules presented in Sections \ref{sec:CaseStudy} and \ref{appendix}, is that such rules are in a one-to-one, straightforward correspondence with the interpretation rules she had in mind, making their formalization easy to address at least in the user-friendly syntax presented in Section \ref{sec:OntoScene_Framework}.

The overall design of our framework allows to easily change both the domain, modifying the ontology in the domain specific parts (under \verb|Classification| and \verb|Interpretation| classes), the used geometric relationships, and the Prolog rules (that are formalized in an external file): furthermore its inclusion in an already existing JADE MAS is quite simple (as described in Section \ref{sec:3.2.4}) thanks to the adoption of the standard JADE usage of OWL ontologies. This makes the exploitation of our framework for other visual languages and existing systems easily achievable.

%The central objective that we set ourselves for this thesis was to develop a system as modular and generic as possible so that every single concept known within it could be easily formalized, modified and replaced (the logical rules of interpretation, the relations geometric, domain ontology and not least the implementations themselves of the concepts) in such a way as to support the end user in the interpretation of complex scenes of various domains. These objectives have been achieved and the architectural choices adopted have proved to be suitable for managing the problem as desired.

The case study presented in Section \ref{sec:CaseStudy} comes from the IndianaMAS project. The results obtained from the experiments are encouraging and demonstrate the flexibility of our approach. The failures that we have reported might have been solved by minor changes to the rules or to the parameters therein. Given that the purpose of our experiments was neither to stress-test the framework, nor to provide a systematic evaluation of its precision and recall, but to show its applicability to a real domain, we left them as hints for a practical use of the framework.

Many improvements can be made to OntoScene. 

So far, we assume that the Detector associates one bounding box with each sub-image: we did not take the possibility of detection ambiguity into account, as we assume that the Detector operates in a deterministic way. Apart from a growing time complexity, there would be no technical obstacles in allowing the Detector to produce more solutions (we mean producing, for the same input image, different decompositions into the sub-images detected there, namely different ``sets of recognized bounding boxes'') and then deal with each of them separately, by running the Classifier and the SceneInterpreter on each of them.

Also, scenes are sensitive to orientation. While this is the correct approach in the rock art domain, where the interpretation may change depending, for example, on one sub-image being above or below another, it might turn out to be a limitation in other domains. 

As far as Prolog rules are concerned, we only used rules meeting a very specific pattern: the initial part of the rule deals with the selection of the scenes to be aggregated, while the second part computes the geometric relations holding among them. This pattern worked well in the rock art domain, but more properties could be associated with images, ranging from features intrinsic to the image itself like the color, to semantically or emotionally related notions like the mood, and these properties could be part of the rules as well. OntoScene allows to add new properties to the \verb|Image| class in the ontology, and use these properties within the logical rules, according to the needs of the end user. For example, we might want to extend the example presented in Figure \ref{fig:19}, and define a happy, red warrior. The Prolog rule might be 
\begin{verbatim}
rule('Red_Happy_Warrior', Scenes) :-
  scene(ID1, Img1, Class1, 'Human', Conf1, SS1),
  scene(ID2, Img2, Class2, 'Sword', Conf2, SS2),
  append([scene(ID1, Img1, Class1, 'Human', Conf1, SS1)],
         [scene(ID2, Img2, Class2, 'Sword', Conf2, SS2)],Scenes), 
  bb(Img1, BB1), 
  bb(Img2, BB2),
  mood(Img1, 'Happy'),
  color(Img2, 'Red'),
  relations(GR),
  jpl_call(GR, overlap, [BB1, BB2], @(true)). 
\end{verbatim}
where \texttt{mood}  and \texttt{color} appear before the \texttt{relations} predicate.

%A post-processing of the \verb|SceneInterpreter| output could be added as a further stage to either filter or rank the interpretations computed by \verb|SceneInterpreter|. 

Another extension we could address in the close future, is to improve geometric relationships. OntoScene supports the addition and definition of new arbitrarily complex geometric relationships: the \verb|Image| class in the ontology can be extended with new geometric properties as the area, the notion of \verb|BB| can be refined by using a polygonal closed line instead of a rectangle, and so on: the framework puts no limits on the type of accepted geometric relationships.

Finally, engraved rock art scenes are represented by black-white, bidimensional images often containing just a few elements placed in relatively simple geometric relationships.
Given that the two phases of the \verb|SceneInterpreter| computation (the creation of the scenes graph and the generation of interpretations) are computationally heavy, they might require optimizations to scale to more complex domains. The possibility to improve the \verb|SceneInterpreter| efficiency by rewriting it in ASP or CLP(FD) is under evaluation, although, before facing this language shift, we should find a domain where scenes are as complex as to motivate it. 

The Prolog code for the \verb|SceneInterpreter| and for some of the examples used for our experiments, and the OWL representation of the ontology, are currently available ``as they are'' from \url{http://www.disi.unige.it/person/MascardiV/Download/OntoScene.zip}. Once the above improvements will be ready, we  plan to make OntoScene available to the research community via a well designed website, after a suitable addition of comments, tutorials, and a user guide in English. 

\section*{Acknowledgments}
We thank Prof. Henry de Lumley and Annie Echassoux for granting us the permission to reproduce some figures from their book \cite{Bego}, and Martine Bert\'ea, Rights Director of CNRS \'editions, for helping us in obtaining their permission.

We are grateful to Dr. Nicoletta Bianchi for her precious support in the IndianaMAS project and in the activities we faced after its conclusion.

Finally, we thank the anonymous reviewers for their thorough reading and for their constructive comments.
\newpage
\bibliographystyle{acmtrans}
\bibliography{Riferimenti}

\newpage

\section{Appendix}
\label{appendix}

\subsection{Ritual sacrifice}
\label{appendix:rs}

\noindent {\bf Association between sub-image classification and sub-image interpretation}:
\begin{scriptsize}
\begin{verbatim}
interpretation('Corniform_Class', 'Corniform').
interpretation('Halberd_Class', 'Halberd').
\end{verbatim}
\end{scriptsize}

\noindent {\bf Rules for scene interpretation}:
\\
\begin{scriptsize}
\begin{verbatim}
rule('Group_Of_Corniforms', Scenes):-...
   % Rule for interpreting a group of corniforms, Section 5.3.
   
rule('Ritual_Sacrifice', Scenes) :-
   scene(ID1, BB1, Class1, 'Halberd', Conf1, SS1),
   (Victim = 'Corniform'; 
    Victim = 'Group_Of_Corniforms'),
   scene(ID2, BB2, Class2, Victim, Conf2, SS2),
   append([scene(ID1, BB1, Class1, 'Halberd', Conf1, SS1)],
          [scene(ID2, BB2, Class2, Victim, Conf2, SS2)], Scenes),
   relations(GR),
   (jpl_call(GR, contains, [BB1, BB2], @(true)); 
    jpl_call(GR, overlap, [BB1, BB2], @(true))).
\end{verbatim}
\end{scriptsize}

\subsection{Bull God birth}
\label{appendix:bgb}
%%%%%%%%%%%%%%%%%%%%%%%%%%%%%%%%%%%%%%%%%%%%%%%%%%%%%%%%%%%

\noindent {\bf Association between sub-image classification and sub-image interpretation}:
\begin{scriptsize}
\begin{verbatim}
interpretation('Double_Appendixes', 'High_Goddess').
interpretation('Corniform_Class', 'Bull_God').
\end{verbatim}
\end{scriptsize}

\noindent {\bf Rules for scene interpretation}:
\begin{scriptsize}
\begin{verbatim}
rule('HG_Giving_Birth_BG', Scenes) :-
  scene(ID1, BB1, Class1, 'High_Goddess', Conf1, SS1),
  scene(ID2, BB2, Class2, 'Bull_God', Conf2, SS2),
  append([scene(ID1, BB1, Class1, 'High_Goddess', Conf1, SS1)], 
         [scene(ID2, BB2, Class2, 'Bull_God', Conf2, SS2)],
         Scenes),
  relations(GR),
  jpl_call(GR, vertical, [BB1, BB2, 'up'], @(true)),
  (jpl_call(GR, near, [BB1, BB2, 0.5], @(true));
   jpl_call(GR, overlap, [BB1, BB2], @(true)))
\end{verbatim}
\end{scriptsize}

\subsection{Storm God}
\label{appendix:sg}

\noindent {\bf Association between sub-image classification and sub-image interpretation}:
\begin{scriptsize}
\begin{verbatim}
interpretation('Dagger_Class', 'Dagger').
interpretation('Reticulum_Class', 'Reticulum').
\end{verbatim}
\end{scriptsize}

\noindent {\bf Rules for scene interpretation}:
\begin{scriptsize}
\begin{verbatim}
rule('Storm_God', Scenes) :-
   scene(ID1, BB1, Class1, 'Dagger', Conf1, SS1),
   scene(ID2, BB2, Class2, 'Reticulum', Conf2, SS2),
   append([scene(ID1, BB1, Class1, 'Dagger', Conf1, SS1)],
          [scene(ID2, BB2, Class2, 'Reticulum', Conf2, SS2)],Scenes),
   relations(GeometricRelations),
   jpl_call(GeometricRelations, overlap, [BB1, BB2], @(true)).
   \end{verbatim}
\end{scriptsize}

\subsection{Rain Invocation}

\noindent {\bf Interpretation of the scene by archeologists}: One human wielding a halberd or an axe (or in general a weapon) represents the rain invocation.
From the analysis of the available images, we discovered that the weapon is usually on above  the human, positioned in a vertical or diagonal way.\\

\noindent {\bf Association between sub-image classification and sub-image interpretation}:
\begin{scriptsize}
\begin{verbatim}
interpretation('Human_Class', 'Human').
interpretation('Halberd_Class', 'Halberd').
interpretation('Axe_Class', 'Axe').
\end{verbatim}
\end{scriptsize}

\noindent {\bf Rules for scene interpretation}:
\begin{scriptsize}
\begin{verbatim}
rule('Rain_Summon', Scenes) :-
   scene(ID1, BB1, Class1, 'Human', Conf1, SS1),
   subclass_of('Weapon_Class', Class),
   interpretation(Class, Weapon),
   scene(ID2, BB2, Class2, Weapon, Conf2, SS2),
   append([scene(ID1, BB1, Class1, 'Human', Conf1, SS1)],
   [scene(ID2, BB2, Class2, Weapon, Conf2, SS2)], Scenes),
   relations(GR),
   jpl_call(GR, vertical, [BB2, BB1, 'up'], @(true)),
   !,
   jpl_call(GR, near, [BB1, BB2, 0.5], @(true)).

rule('Rain_Summon', Scenes) :-
   [omissis] % as in the previous rule
   relations(GR),
   (jpl_call(GR, diagonal, [BB2, BB1, 'ne'], @(true)) ; 
    jpl_call(GR, diagonal, [BB2, BB1, 'nw'], @(true))),
    jpl_call(GR, near, [BB1, BB2, 0.5], @(true)).
\end{verbatim}
\end{scriptsize}

\noindent {\bf Explanation}: the rule searches for a human figure, then it searches for a weapon (note that halberd and axe are subclasses of weapon in the domain ontology, so we write a general rule including all the weapons as required by the archeologists) and checks for the correct geometrical relationship; then, the rule checks if the BBs of the human and of the weapon are close to each other, and if the one of the weapon is above the human, in vertical or diagonal relationship.\\

\noindent Table \ref{tab:rain} reports the results of the four analyzed images, all correctly interpreted. %: the last one is not identified because the two bounding boxes are not overlapping neither one inside the other, as foreseen in the rule.

\begin{table}%[!htb]
\begin{tiny}
\begin{tabular}{m{1.6cm}m{4.5cm}m{6cm}m{0.8cm}}
\toprule
Image & Input Single Images & Resulting Interpretation & Final result\\
\midrule
\includegraphics[width=1.3cm]{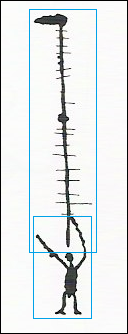} DE, P. 189, F. 157(4) & 

\verb|image(0,[('Human_Class', 1.0)])|

\verb|image(1,[('Halberd_Class',1.0)])| & 

\verb|I1=[Rain_Summon(Human-0,Halberd-1)]|

\verb|I2=[Human-0, Halberd-1]|& Passed\\

\includegraphics[width=1.3cm]{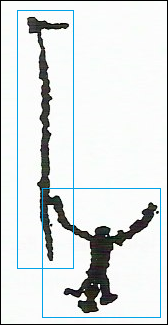} DE, P. 189, F. 157(6)  & 

\verb|image(0,[('Human_Class', 1.0)])|

\verb|image(1,[('Halberd_Class',1.0)])| & 

\verb|I1=[Rain_Summon(Human-0,Halberd-1)]|

\verb|I2=[Human-0, Halberd-1]|& Passed\\

\includegraphics[width=1.3cm]{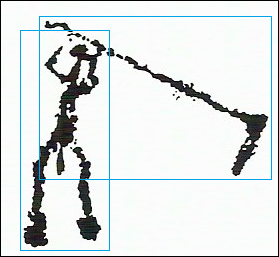} DE, P. 189, F. 157(12)  &

\verb|image(0,[('Human_Class', 1.0)])|

\verb|image(1,[('Halberd_Class',1.0)])| & 

\verb|I1=[Rain_Summon(Human-0,Halberd-1)]|

\verb|I2=[Human-0, Halberd-1]|& Passed\\

\includegraphics[width=1.3cm]{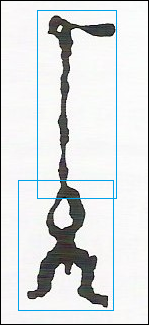} DE, P. 200, F. 171(2)  & 

\verb|image(0,[('Human_Class', 1.0)])|

\verb|image(1,[('Axe_Class',1.0)])| & 

\verb|I1=[Rain_Summon(Human-0,Axe-1)]|

\verb|I2=[Human-0, Axe-1]|& Passed\\
\bottomrule

\end{tabular}
\caption{Results of the interpretation of Rain Invocation scenes.}
\label{tab:rain}
\end{tiny}
\end{table}

%\newpage

\subsection{Queens Fight}

\begin{table}%[!htb]
\begin{tiny}
\begin{tabular}{m{1.5cm}m{4.8cm}m{5.7cm}m{0.8cm}}
\toprule
Image & Input Single Images & Resulting Interpretation & Final result\\
\midrule
\includegraphics[width=1.3cm]{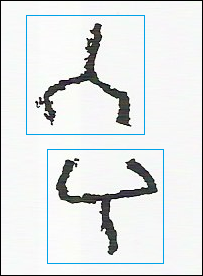} DE, P. 221, F. 191(5) & 

\verb|image(0,[('Up_Down_Corn_Class',1.0)])|

\verb|image(1,[('Up_Corn_Class',1.0)])| & 

\verb|I1=[Queens_Fight(Corniform-0,Corniform-1)]|

\verb|I2=[Corniform-0,Corniform-1]|& Passed\\

\includegraphics[width=1.3cm]{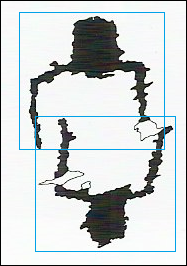} DE, P. 221, F. 191(7)  & 

\verb|image(0,[('Up_Down_Corn_Class',1.0)])|

\verb|image(1,[('Up_Corn_Class',1.0)])| & 

\verb|I1=[Queens_Fight(Corniform-0,Corniform-1)]|

\verb|I2=[Corniform-0,Corniform-1]|& Passed\\

\includegraphics[width=1.3cm]{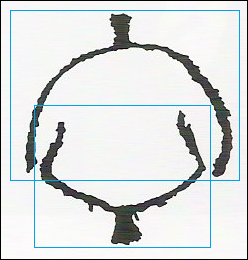} DE, P. 221, F. 191(8)  &

\verb|image(0,[('Up_Down_Corn_Class',1.0)])|

\verb|image(1,[('Up_Corn_Class',1.0)])| & 

\verb|I1=[Queens_Fight(Corniform-0,Corniform-1)]|

\verb|I2=[Corniform-0,Corniform-1]|& Passed\\

\includegraphics[width=1.3cm]{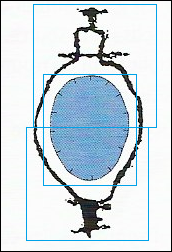} DE, P. 221, F. 191(9)  & 

\verb|image(0,[('Up_Down_Corn_Class',1.0)])|

\verb|image(1,[('Up_Corn_Class',1.0)])| 

\verb|image(2,[('Rock_Class',1.0)])| & 

\verb|I1=[Corniform-0,Corniform-1, Rock-1]|& Failed (Around not available)\\

\bottomrule

\end{tabular}
\caption{Results of the interpretation of Queens Figth.}
\label{tab:queens}
\end{tiny}
\end{table}

\noindent {\bf Interpretation of the scene by archeologists}: Two corniforms with juxtaposed horns represent a ritual fighting called in archeology ``the Queens Fight''.
The two corniforms must be one over the other, with contrary directions of the horns (we assume that the classifier is able to discriminate between the two different positions), and their BBs may, or may not, intersect, but should be close to each other.\\

\noindent {\bf Association between sub-image classification and sub-image interpretation}:
\begin{scriptsize}
\begin{verbatim}
interpretation('Up_Corn_Class', 'Corniform').
interpretation('Up_Down_Corn_Class', 'Corniform').
\end{verbatim}
\end{scriptsize}

\noindent {\bf Rules for scene interpretation}:
\begin{scriptsize}
\begin{verbatim}
rule('Queens_Fight', Scenes) :-
   scene(ID1, BB1, 'Up_Down_Corn_Class', 'Corniform',Conf1, SS1),
   scene(ID2, BB2, 'Up_Corn_Class', 'Corniform',Conf2, SS2),
   append([scene(ID1, BB1, 'Up_Down_Corn_Class','Corniform', Conf1, SS1)],
          [scene(ID2, BB2, 'Up_Corn_Class','Corniform', Conf2, SS2)], Scenes),
   relations(GR),
   jpl_call(GR, vertical, [BB1, BB2, 'up'], @(true)),
   jpl_call(GR, near, [BB1, BB2, 0.5], @(true)).
\end{verbatim}
\end{scriptsize}

\noindent {\bf Explanation}: the rule searches for two corniforms, one with up horns and the other with down horns, in vertical relationship and close to each other. \\

\noindent Table \ref{tab:queens} reports the results of the four analyzed images: the last one is not correctly interpreted because the geometrical relationships ``Around'' has not been implemented yet, and a third unexpected element (a rock) appears in the scene.

%\newpage

\subsection{Bull God}

\noindent {\bf Interpretation of the scene by archeologists}: One corniform inside the horns of another one represents the Bull God.
By analyzing the available images, two patterns were discovered: the first is one or more corniforms inside another one, another is a group of corniforms vertically aligned, not necessary one inside the other.\\

\noindent {\bf Association between sub-image classification and sub-image interpretation}:
\begin{scriptsize}
\begin{verbatim}
interpretation('Up_Corn_Class', 'Corniform').
\end{verbatim}
\end{scriptsize}

\noindent {\bf Rules for scene interpretation}:
\begin{scriptsize}
\begin{verbatim}
rule('Group_Of_Corniforms', Scenes) :- .... (from Section 5.3)

rule('Bull_God', Scenes) :- (Inner = 'Corniform'; 
   Inner = 'Group_Of_Corniforms'),
   scene(ID1, BB1, Class1, Inner, Conf1, SS1),
   scene(ID2, BB2, Class2, 'Corniform', Conf2, SS2),
   append([scene(ID1, BB1, Class1, Inner, Conf1, SS1)],
          [scene(ID2, BB2, Class2, 'Corniform', Conf2, SS2)], Scenes),
   relations(GR),
   jpl_call(GR, contains, [BB2, BB1], @(true)).

rule('Bull_God', Scenes) :-
   get_corniforms_same_direction(Corniforms), 
   sublist(Corniforms, Scenes), length(Scenes, Len), Len > 1,
   findall(BB, member(scene(_, BB, _, _, _, _), Scenes), BBs),
   relations(GR),
   test_vertical(BBs, GR).
\end{verbatim}
\end{scriptsize}

\noindent {\bf Explanation}: the first rule selects one corniform (or a group of corniforms) and another one from the list, and checks if they are one inside the other. The second rule uses the predicate \verb|get_corniforms_same_direction| to get all the corniforms with the same orientation and checks if they are in vertical relationships. We omit here the definition of the \texttt{test\_ver\-ti\-cal(BBs, GR)} predicate.\\

\noindent Table \ref{tab:bullgod} reports the results of the four analyzed images, all correctly interpreted. %: the last one is not identified because the geometrical relationships Around is not yes available, and further more a third element (a rock) appears in the figure.

\begin{table}[!htb]
\begin{tiny}
\begin{tabular}{m{1.5cm}m{4.2cm}m{6.5cm}m{0.5cm}}
\toprule
Image & Input Single Images & Resulting Interpretation & Final result\\
\midrule
\includegraphics[width=1.3cm]{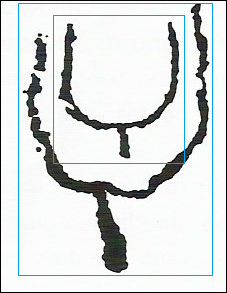} DE, P. 218, F. 186(2) & 

\verb|image(0,[('Up_Corn_Class',1.0)])|

\verb|image(1,[('Up_Corn_Class',1.0)])| & 

\verb|I1=[Bull_God(Corn-0,Corn-1)]|

\verb|I2=[Corn-0,Corn-1]|& Passed\\

\includegraphics[width=1.3cm]{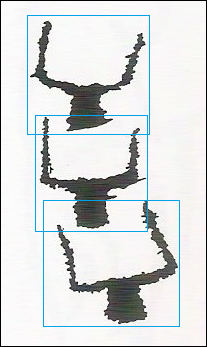} DE, P. 218, F. 186(9)  & 

\verb|image(0,[('Up_Corn_Class',1.0)])|

\verb|image(1,[('Up_Corn_Class',1.0)])|

\verb|image(2,[('Up_Corn_Class',1.0)])| & 

\verb|I1=[Bull_God(Corn-0,Corn-1,Corn-2)]|

\verb|I2=[Corn-0,Corn-1,Corn-2]|& Passed\\

\includegraphics[width=1.3cm]{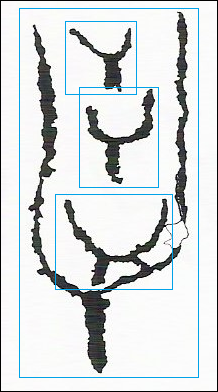} DE, P. 218, F. 186(7)  &

\verb|image(0,[('Up_Corn_Class',1.0)])|

\verb|image(1,[('Up_Corn_Class',1.0)])|

\verb|image(2,[('Up_Corn_Class',1.0)])| 

\verb|image(3,[('Up_Corn_Class',1.0)])| & 

\verb|I1=[Bull_God(Corn-0,Corn-1,Corn-2,Corn-3)]|

\verb|I2=[Corn-0, Bull_God(Corn-1,Corn-2,Corn-3)]|

\verb|I3=[Corn-0,Corn-3,Bull_God(Corn-1, Corn-2)]| 

\verb|I4=[Corn-0,Corn-1,Corn-2,Corn-3)]| & Passed\\

\includegraphics[width=1.3cm]{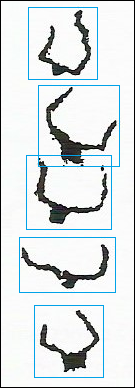} DE, P. 218, F. 186(18)  & 

\verb|image(0,[('Up_Corn_Class',1.0)])|

\verb|image(1,[('Up_Corn_Class',1.0)])|

\verb|image(2,[('Up_Corn_Class',1.0)])| 

\verb|image(3,[('Up_Corn_Class',1.0)])|

\verb|image(4,[('Up_Corn_Class',1.0)])| & 

\verb|I1=[Bull_God(Corn-0,Corn-1,Corn-2,Corn-3, Corn-4)]|

\verb|I2=[Corn-0,Bull_God(Corn-1, Corn-2,Corn-3,Corn-4)]| 

\verb|I3=[Bull_God(Corn-0, Corn-1,Corn-2),|

\verb|Bull_God(Corn-3, Corn-4)]| 

\verb|... other 7 interpretations| & Passed\\

\bottomrule

\end{tabular}
\caption{Results of the interpretation of Bull God.}
\label{tab:bullgod}
\end{tiny}
\end{table}

%\newpage

%%%%%%%%%%%%%%%%%%%%%%%%%%%%%%%%%%%%%%%%%%%%%%%%%%%%%%%%%%%

\subsection{Rain Propitiatory Rite}

\noindent {\bf Interpretation of the scene by archeologists}: One dagger between the horns of a corniform represents a propitiatory rite for the rain. The two sub-images should intersect and at the same time the dagger should be partially inside the horns, above them. With the currently implemented geometrical relationships we cannot express this relation in a precise way, so we approximated it.\\

\noindent {\bf Association between sub-image classification and sub-image interpretation}:
\begin{scriptsize}
\begin{verbatim}
interpretation('Corniform_Class', 'Corniform').
interpretation('Dagger_Class', 'Dagger').
\end{verbatim}
\end{scriptsize}

\noindent {\bf Rules for scene interpretation}:
\begin{scriptsize}
\begin{verbatim}
rule('Rain_Propitiatory_Rite', Scenes) :-
   scene(ID1, BB1, Class1, 'Dagger', Conf1, SS1),
   scene(ID2, BB2, Class2, 'Corniform', Conf2, SS2),
   append([scene(ID1, BB1, Class1, 'Dagger', Conf1, SS1)],
          [scene(ID2, BB2, Class2, 'Corniform', Conf2, SS2)], Scenes),
   relations(GR),
   jpl_call(GR, vertical, [BB1, BB2, 'up'], @(true)),
   jpl_call(GR, overlap, [BB1, BB2], @(true)).
\end{verbatim}
\end{scriptsize}

\noindent {\bf Explanation}: the rule searches for a dagger and a corniform, checking if they overlap and if the dagger is above the corniform.\\

\noindent Table \ref{tab:rain-propitiatory} reports the results of the three analyzed images, all correctly interpreted. %: the last one is not identified because the geometrical relationships Around is not yes available, and further more a third element (a rock) appears in the figure.

\begin{table}[!htb]
\begin{tiny}
\begin{tabular}{m{1.5cm}m{4.8cm}m{5.7cm}m{0.8cm}}
\toprule
Image & Input Single Images & Resulting Interpretation & Final result\\
\midrule
\includegraphics[width=1.3cm]{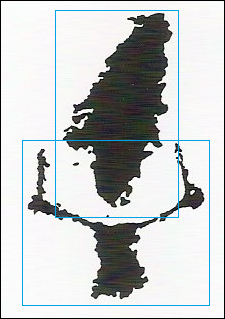} DE, P. 170, F. 132(1) & 

\verb|image(0,[('Dagger_Class',1.0)])|

\verb|image(1,[('Corniform_Class',1.0)])| & 

\verb|I1=[Rain_Propitiatory_Rite(Dagger-0, Corniform-1)]|

\verb|I2=[Dagger-0, Corniform-1]|& Passed\\

\includegraphics[width=1.3cm]{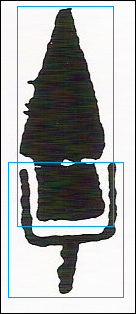} DE, P. 170, F. 132(2)  & 

\verb|image(0,[('Dagger_Class',1.0)])|

\verb|image(1,[('Corniform_Class',1.0)])| & 

\verb|I1=[Rain_Propitiatory_Rite(Dagger-0, Corniform-1)]|

\verb|I2=[Dagger-0, Corniform-1]|& Passed\\

\includegraphics[width=1.3cm]{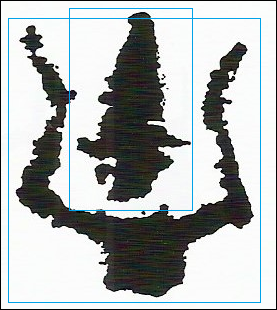} DE, P. 170, F. 132(7)  &

\verb|image(0,[('Dagger_Class',1.0)])|

\verb|image(1,[('Corniform_Class',1.0)])| & 

\verb|I1=[Rain_Propitiatory_Rite(Dagger-0, Corniform-1)]|

\verb|I2=[Dagger-0, Corniform-1]|& Passed\\

\bottomrule

\end{tabular}
\caption{Results of the interpretation of Rain Propitiatory Rite.}
\label{tab:rain-propitiatory}
\end{tiny}
\end{table}

%\newpage

\subsection{Agricultural Rite}
\label{appendix:arit}

\noindent {\bf Interpretation of the scene by archeologists}: One or two priests making water spring from an artificial repository represent an agricultural rite. The most recurring pattern includes one or two humans holding a repository, which is above them, from which the water falls down.\\

\noindent {\bf Association between sub-image classification and sub-image interpretation}:
\begin{scriptsize}
\begin{verbatim}
interpretation('Human_Class', 'Priest').
interpretation('Repository_Class', 'Repository').
interpretation('Water_Class', 'Water').
\end{verbatim}
\end{scriptsize}

\noindent {\bf Rules for scene interpretation}:
\begin{scriptsize}
\begin{verbatim}
rule('Agricultural_Rite',
   [scene(ID1, BB1, Class1, 'Priest', Conf1, SS1),
    scene(ID2, BB2, Class2, 'Priest', Conf2, SS2),
    scene(ID3, BB3, Class3, 'Repository', Conf3, SS3),
    scene(ID4, BB4, Class4, 'Water', Conf4, SS4)]) :-
  scene(ID1, BB1, Class1, 'Priest', Conf1, SS1),
  scene(ID2, BB2, Class2, 'Priest', Conf2, SS2),
  scene(ID3, BB3, Class3, 'Repository', Conf3, SS3),
  scene(ID4, BB4, Class4, 'Water', Conf4, SS4),
  relations(GR),
  jpl_call(GR, diagonal, [BB1, BB3, 'sw'], @(true)),
  jpl_call(GR, near, [BB1, BB3, 0.5], @(true)),
  jpl_call(GR, diagonal, [BB2, BB3, 'se'], @(true)),
  jpl_call(GR, near, [BB2, BB3, 0.5], @(true)),
  jpl_call(GR, vertical, [BB3, BB4, 'up'], @(true)),
  jpl_call(GR, near, [BB3, BB4, 0.5], @(true)).\end{verbatim}
\end{scriptsize}

\noindent {\bf Explanation}: the rule searches for the two humans, the water and the repository, checking if the two humans are in diagonal (one on the left and one on the right) below the repository, and if the water is under the repository. All the images should be close to each other. Another rule, searching for only one human, is not reported since it is very similar to one shown here.\\

\noindent Table \ref{tab:agricultural} reports the results of the three analyzed images: the last one has not been correctly interpreted because the human is above (not below and in diagonal) the repository.

\begin{table}[!htb]
\begin{tiny}
\begin{tabular}{m{1.5cm}m{4.8cm}m{5.7cm}m{0.8cm}}
\toprule
Image & Input Single Images & Resulting Interpretation & Final result\\
\midrule
\includegraphics[width=1.3cm]{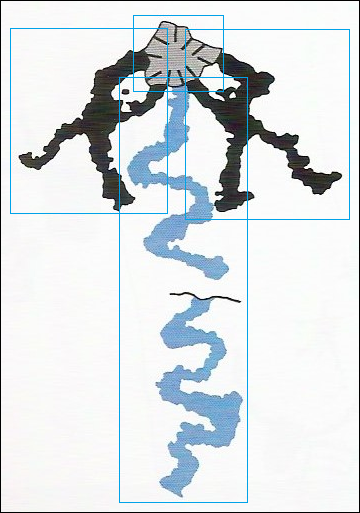} DE, P. 284, F. 268(1) & 

\verb|image(0,[('Human_Class',1.0)])|

\verb|image(1,[('Human_Class',1.0)])| 

\verb|image(2,[('Repository_Class',1.0)])| 

\verb|image(3,[('Water_Class',1.0)])| & 

\verb|I1=[Agricultural_Rite(Priest-0, Priest-1,|

\verb|Repository-2, Water-3)]|

\verb|I2=[Priest-0,Priest-1,Repository-2,Water-3]|& Passed\\

\includegraphics[width=1.3cm]{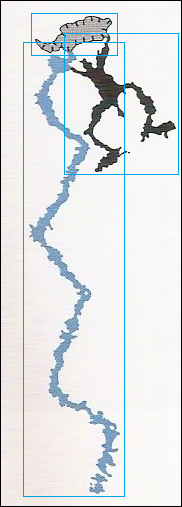} DE, P. 284, F. 268(4)  & 

\verb|image(0,[('Human_Class',1.0)])|

\verb|image(1,[('Repository_Class',1.0)])| 

\verb|image(2,[('Water_Class',1.0)])| & 

\verb|I1=[Agricultural_Rite(Priest-0,|

\verb|Repository-1, Water-2)]|

\verb|I2=[Priest-0,Repository-1,Water-2]|& Passed\\

\includegraphics[width=1.3cm]{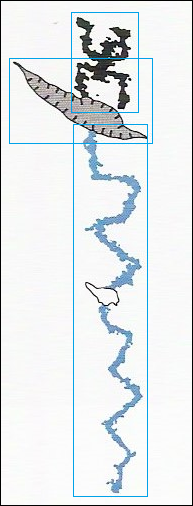} DE, P. 284, F. 268(5)  &

\verb|image(0,[('Human_Class',1.0)])|

\verb|image(1,[('Repository_Class',1.0)])| 

\verb|image(2,[('Water_Class',1.0)])| & 

\verb|I1=[Human-0,Repository-1,Water-2]|& Failed (Human over repository)\\

\bottomrule

\end{tabular}
\caption{Results of the interpretation of Agricultural Rite.}
\label{tab:agricultural}
\end{tiny}
\end{table}

\end{document}